\documentclass[a4paper, 11pt, abstracton, oneside]{scrartcl}  
\usepackage[authoryear,sort&compress]{natbib}
\usepackage[a4paper,top=2.0cm,bottom=2.0cm,left=2.0cm,right=2.0cm]{geometry}
\usepackage[utf8x]{inputenc}
\usepackage[english]{babel}
\usepackage{graphicx, amsfonts,amsmath,amssymb,amsthm,bm}
\usepackage{commands}
\usepackage{longtable,fancybox}
\usepackage{comment,geometry}
\usepackage{booktabs,multirow}
\usepackage{makecell}
\usepackage{threeparttable}
\usepackage{caption}
\usepackage{enumitem}
\usepackage{multicol}
\usepackage{placeins}
\usepackage{glossaries}
\glsdisablehyper

\usetikzlibrary{positioning,calc,arrows.meta,fit,shapes.misc,decorations.pathreplacing}
\usepackage{algorithm,algpseudocode}

\newacronym{acr:ml}{ML}{machine learning}
\newacronym{acr:or}{OR}{operations research}
\newacronym{acr:cso}{CSO}{contextual stochastic optimization}
\newacronym{acr:coaml}{COAML}{combinatorial optimization augmented ML}
\newacronym{acr:ecm}{ECM}{empirical cost minimization}
\newacronym{acr:co}{CO}{combinatorial optimization}
\newacronym{acr:rl}{RL}{reinforcement learning}
\newacronym{acr:saa}{SAA}{sample average approximation}
\newacronym{acr:pto}{PTO}{predict-then-optimize}
\newacronym{acr:mse}{MSE}{mean squared error}
\newacronym{acr:srl}{SRL}{structured RL}
\newacronym{acr:fy}{FY}{Fenchel-Young}
\newacronym{acr:il}{IL}{imitation learning}
\newacronym{acr:mlp}{MLP}{multi-layer perceptron}
\newacronym{acr:nn}{NN}{neural network} 
\newacronym{acr:gnn}{GNN}{graph neural network}
\newacronym{acr:glm}{GLM}{generalized linear model}
\newacronym{acr:pinn}{PINN}{physics-informed NN}
\newacronym{acr:sl}{SL}{supervised learning}
\newacronym{acr:tsp}{TSP}{traveling salesman problem}
\newacronym{acr:vrp}{VRP}{vehicle routing problem}
\newacronym{acr:dvrp}{DVRP}{dynamic VRP}
\newacronym{acr:dirp}{DIRP}{dynamic IRP}
\newacronym{acr:cvrp}{CVRP}{capacitated VRP}
\newacronym{acr:vsp}{VSP}{vehicle scheduling problem}
\newacronym{acr:svsp}{SVSP}{stochastic VSP}
\newacronym{acr:smsp}{SMSP}{single machine scheduling problem}
\newacronym{acr:fctp}{FCTP}{fixed charge transportation problem}
\newacronym{acr:cnn}{CNN}{convolutional neural network}
\newacronym{acr:milp}{MILP}{mixed-integer linear programming}
\newacronym{acr:mdp}{MDP}{Markov decision process}
\newacronym{acr:rnn}{RNN}{recurrent neural network}

\usepackage{etoolbox}  

\newlength{\myglsitemsep}
\newglossarystyle{acronymsupercompact}{%
  \setglossarystyle{list}%
    {\begin{description}[
        itemsep=0pt,
        parsep=0pt,
        topsep=0pt,
        partopsep=0pt,
        leftmargin=1.5em
      ]}%
    {\end{description}}%
}

\usepackage[mode=buildnew]{standalone}
\usepackage{authblk}

\RedeclareSectionCommand[
  beforeskip=6pt
]{paragraph}

\title{\textbf{Combinatorial Optimization Augmented Machine Learning}\\
\vspace{0.3em}\large Technical Foundations, Applications, and Research Frontiers}

\author[1,2]{Maximilian Schiffer}
\author[1,2]{Heiko Hoppe}
\author[3]{Yue Su}
\author[4]{Louis Bouvier}
\author[4]{Axel Parmentier}

\affil[1]{\small School of Management, Technical University of Munich, Germany}
\affil[2]{\small Munich Data Science Institute, Technical University of Munich, Germany}
\affil[3]{Inria, Université de Lille, France}
\affil[4]{CERMICS, Ecole Nationale des Ponts et Chaussées, Institut Polytechnique Paris, France}

\makenoidxglossaries
\begin{document}
\maketitle

\begin{abstract}
Combinatorial optimization augmented machine learning (COAML) has recently emerged as a powerful paradigm for integrating predictive models with combinatorial decision-making. 
By embedding combinatorial optimization oracles into learning pipelines, COAML enables the construction of policies that are both data-driven and feasibility-preserving, bridging the traditions of machine learning, operations research, and stochastic optimization. 
This paper provides a comprehensive overview of the state of the art in COAML. We introduce a unifying framework for COAML pipelines, describe their methodological building blocks, and formalize their connection to empirical cost minimization. 
We then develop a taxonomy of problem settings based on the form of uncertainty and decision structure. 
Using this taxonomy, we review algorithmic approaches for static and dynamic problems, survey applications across domains such as scheduling, vehicle routing, stochastic programming, and reinforcement learning, and synthesize methodological contributions in terms of empirical cost minimization, imitation learning, and reinforcement learning.
 Finally, we identify key research frontiers. 
This survey aims to serve both as a tutorial introduction to the field and as a roadmap for future research at the interface of combinatorial optimization and machine learning.
\end{abstract}
\vspace{1em}

\section{Introduction}
\begin{figure}[!ht]
	\centering
	\includegraphics[width=\textwidth]{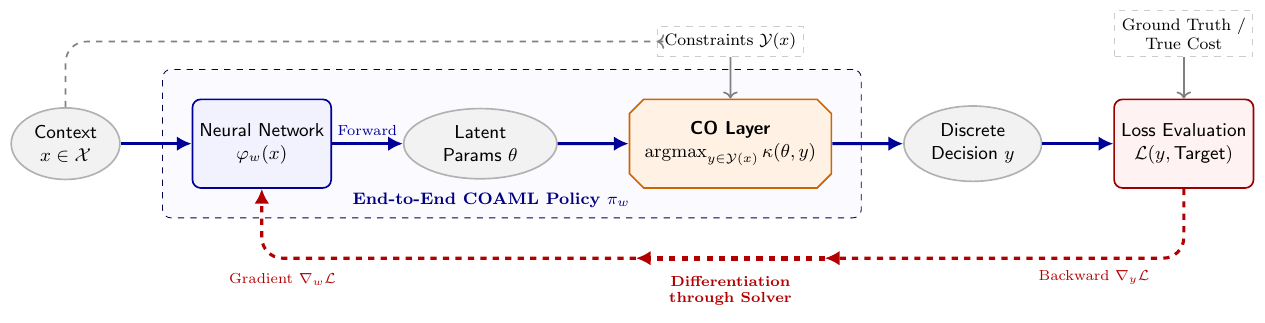}
	\caption{Conceptual illustration of the \gls{acr:coaml} paradigm. A \gls{acr:nn} parameterizes a surrogate \gls{acr:co} problem. We propagate gradients backwards through the \gls{acr:co} layer, enabling end-to-end training of the policy $\pi_w$ to minimize a downstream decision loss.}
	\label{fig:coamlSketch}
\end{figure}

The convergence of \gls{acr:ml} and \gls{acr:or} stands as a central development in modern decision science. While \gls{acr:ml} excels at extracting patterns from high-dimensional data to forecast uncertain parameters, \gls{acr:or} provides the rigorous machinery to navigate complex, discrete feasible regions and identify optimal decisions. Traditionally, these capabilities have been deployed sequentially—predicting parameters first, then optimizing deterministic models. However, this separation often fails to align predictive accuracy with downstream decision quality. \Gls{acr:coaml} addresses this gap by integrating combinatorial algorithms directly into the learning loop, creating end-to-end architectures that learn to optimize, see Figure~\ref{fig:coamlSketch}. To formalize this paradigm and its necessity, we begin by examining the standard framework of \gls{acr:cso}, as it yields a natural starting point to data-driven approaches in \gls{acr:or}.

Consider a setting where a decision maker observes a context $x \in \calX$, containing all available information required to make a decision $y$. This decision must belong to a (combinatorial) set of feasible decisions~$\calY(x)$. Once the decision is implemented, an uncertainty $\xi$ realizes, incurring a cost $c(x,y,\xi)$.
We assume a joint but unknown probability distribution $\bbP$ over $(\bfx,\bfxi)$, with $\bfx,\bfxi$ being random variables.
Throughout this survey, we adhere to this convention, denoting random variables by bold symbols, e.g., $\bfx$, and using regular symbols, e.g., $x$, for their realizations.

Our objective is to determine a policy $\pi$ that maps contexts to decisions. To accommodate non-deterministic policies, we generally define $\pi(\cdot|x)$ as a conditional distribution over $y$ given $x$.
Since the decision maker cannot observe $\bfxi$ beforehand, the decision $\bfy$ is conditionally independent of $\bfxi$ given $\bfx$.
In this setting, our goal is to minimize the expected cost
\begin{equation}\label{eq:introCSO}
 \min_\pi \bbE_{(\bfx,\bfxi)\sim \bbP, \, \bfy\sim \pi(\cdot|\bfx)}\Big[c(\bfx,\bfy,\bfxi)\Big].
\end{equation}
An optimal policy maps a context $x$ to a decision $y \in \calY(x)$ that solves the stochastic optimization problem defined by the conditional distribution of $\xi$ given $x$:
\begin{equation}\label{eq:introHarProblemConditionalExpecation}
x \longmapsto y \in \argmin_{y\in \calY(x)} \bbE\Big[c(x,y,\bfxi)\Big|x\Big].
\end{equation}
However, deriving such a policy is generally intractable: statistically, because the distribution over $(\bfx,\bfxi)$ is unknown, and computationally, because the resulting problem is both stochastic and combinatorial.

In the \gls{acr:coaml} framework, we address this intractability by focusing on policies of the form
\begin{equation}\label{eq:introSurrogateProblem}
 \pi_w : x \longmapsto \hat y(\theta) \in \argmax \kappa(\theta, y) \quad \text{where} \quad \theta = \varphi_w(x).
\end{equation}
Here, $\max \kappa(\theta, y)$ represents a \emph{surrogate \gls{acr:co}} problem parameterized by $\theta$, and $\varphi_w$ is a statistical model parameterized by $w \in \calW$, typically a \gls{acr:nn}.
In many cases, $\kappa(\theta, y)$ takes the form of a linear function, such as $\theta^\top y$ or $\theta^\top \phi(y)$ where $\phi(y)$ is an embedding. We assume the oracle $\hat y$ to be efficient and scalable, representing any standard \gls{acr:co} solver—such as mixed-integer programming, shortest-path algorithms, or sorting algorithms. These solvers leverage decades of advancements in mathematical programming, making them broadly applicable across problems.
Note that while this general formulation allows for stochastic policies, we describe $\pi_w$ as a deterministic mapping for the sake of clarity in this introduction. We will generalize this surrogate problem to define a distribution $\pi_w(\cdot|x)$ over $\calY(x)$ in later sections.

Our learning problem is to find parameters $w$ that yield a high-quality policy, solving:
\begin{equation}
    \min_{w\in \calW}  \bbE_{(\bfx,\bfxi)\sim \bbP}\Big[c(\bfx,\pi_w(\bfx),\bfxi)\Big].
\end{equation}
In practice, we lack access to $\bbP$ and must rely on a training set.
The nature of this training set depends on the application, dictating the specific learning methodology.
The simplest setting in \gls{acr:cso} assumes a training set $(x_1,\xi_1),\ldots,(x_n,\xi_n)$ of historical contexts and uncertainties. Full access to $\xi_i$ is common in internal operations optimization, though less guaranteed in settings like sales.
Here, it is natural to frame learning as \emph{\gls{acr:ecm}}:
\begin{equation}
    \label{eq:IntroEmpiricalCostMinimization}
    \min_w \frac{1}{n}\sum_{i=1}^n c(x_i,\pi_w(x_i),\xi_i).
\end{equation}
In some applications, we do not observe $\xi_i$ directly but only the cost $c(x_i,\pi_w(x_i),y_i)$, which may result from applying the policy in the real world or within a simulator. In such cases, the training set consists only of $x_1,\ldots,x_n$, along with the ability to sample from $c(x_i,\pi_w(x_i),\bfxi)$, where $\bfxi$ is distributed according to $\bbP(\bfxi|x_i)$. This limited information setting is referred to as \emph{single-step \gls{acr:rl}} in the context of \gls{acr:ecm}.

In other applications, an expert policy may provide target decisions $(x_1,\bar y_1),\ldots,(x_n,\bar y_n)$ for imitation.
We then formulate the problem as \emph{\gls{acr:sl}}:
\begin{equation}
    \label{eq:IntroDecisionFocusedSupervisedLearning}
    \min_w \frac{1}{n}\sum_{i=1}^n \calL\big(\pi_w(x_i),\bar y_i\big),
\end{equation}
where $\calL(y,\bar y)$ quantifies the divergence between $y$ and $\bar y$.

While other learning settings exist, a key  of the ones presented above is that they are \emph{decision-focused}: the learning objective evaluates the quality of the solution $y$ derived from $\theta$.
This contrasts with \emph{decision-agnostic} approaches, which typically minimize a prediction loss over a training set $(x_1,\theta_1),\ldots,(x_n,\theta_n)$:
\begin{equation}\label{eq:IntroNonDecisionFocusedSupervisedLearning}
    \min_{w}\frac1n\sum_{i=1}^n \ell(\varphi_w(x_i),\theta_i).
\end{equation}
Decision-agnostic learning ignores that the ultimate goal is not maximizing the accuracy of $\theta$, the output of $\varphi_w$, but rather the quality of the decision $ y \in \argmax \kappa(\theta, y)$ for the original problem~\eqref{eq:introHarProblemConditionalExpecation}.

Decision-focused learning is critical for \gls{acr:coaml}'s performance but introduces a practical challenge: the surrogate problem~\eqref{eq:introSurrogateProblem} acts as the final \emph{layer} of the \gls{acr:nn}.
Since \glspl{acr:nn} are trained via first-order methods, a significant portion of \gls{acr:coaml} methodology focuses on making differentiation through \emph{\gls{acr:co} layers} both meaningful and practical.

\subsection{The Imperative of \gls{acr:coaml} in Modern \gls{acr:or}}
The emergence of \gls{acr:coaml} represents a fundamental evolution in how \gls{acr:or} approaches the integration of data and decisions. While the \gls{acr:cso} framework introduced above provides a rigorous mathematical basis for this paradigm, the practical imperative for \gls{acr:coaml} arises from systemic limitations in traditional decision-making architectures. Modern industrial applications demand solutions that are not only performant in theory but also robust to high-dimensional uncertainty and scalable to real-world instances. Conventional methods, which typically segregate prediction from optimization, increasingly struggle to meet these dual requirements. In the following, we examine the structural pressures—ranging from computational scalability to the complexity of stochastic environments—that motivates the shift toward end-to-end learning policies.

\paragraph{Scalability in \gls{acr:co}.}
\gls{acr:or} optimizes industrial processes through efficient resource allocation. Often, resources are indivisible—e.g., a machine is either assigned to a job or not—giving rise to \gls{acr:co} problems. Scaling these problems is essential to unlocking flexibility and economies of scale. For instance, in bin packing, larger volumes allow for more efficient container usage, reducing marginal costs. Similarly, in warehouse order picking, consolidating orders optimizes routes and reduces costs. Realizing these benefits requires algorithms capable of solving large-scale instances efficiently.

\paragraph{The complexity of uncertainty.}
Modern decision-making is characterized by pervasive uncertainty. Disruptions like the COVID-19 pandemic or the variability of renewable energy highlight the necessity of accounting for uncertainty in planning.
Historically, \gls{acr:or} has addressed these challenges independently, employing \gls{acr:co} for static NP-hard problems, stochastic/robust optimization for strategic uncertainty, and control-theoretic methods for sequential decisions. This division frequently requires simplifying assumptions. For instance, navigating vast combinatorial spaces typically depends on simplified objectives, which are common in \gls{acr:milp}. While \gls{acr:milp} solvers have achieved massive speedups (over $500$ billion times between 1991 and 2015), stochastic and robust optimization techniques remain computationally intractable for large-scale real-world applications like supply chain planning or vehicle routing.

\paragraph{Limitations of separated architectures.}
The scalability issues of classic uncertainty-handling methods hinder their broad adoption. Since \gls{acr:or} gains often stem from economies of scale, restricting optimization to small instances to accommodate stochastic complexity is counterproductive. Consequently, practitioners often favor \gls{acr:pto} architectures, where statistical models first generate deterministic approximations, e.g., expected values, which are then solved by mature combinatorial algorithms. While computationally efficient, this separation creates a critical misalignment: the statistical model is typically trained to minimize a symmetric prediction loss, e.g., the \gls{acr:mse}, without awareness of the downstream optimization landscape. For example, consider a \gls{acr:vrp} that bases on predicted customer demand. For a standard regression model, a $5\%$ underestimation of demand is equivalent to a $5\%$ overestimation. However, in practice, underestimating demand may result in a truck arriving without sufficient capacity, necessitating a highly costly recourse action—such as dispatching a second vehicle solely for the overflow. In contrast, a slight overestimation merely results in minor unutilized space. By decoupling prediction accuracy from decision quality, a standard prediction fails to penalize the specific errors that trigger such expensive operational repairs.

\paragraph{The methodological pivot to structured learning.}
The \gls{acr:ml} community encountered analogous challenges in the early 2000s with \emph{structured output prediction} tasks, such as image segmentation or natural language parsing. Unlike simple regression or classification where predictions are independent scalars, these tasks require predicting a complex configuration of interdependent variables, effectively forming a combinatorial object. Initially, the dominant architecture treated prediction and optimization separately. The standard approach was to first estimate the full joint probability distribution $\bbP(\bfxi|\bfx)$ and then solve an optimization problem—often termed inference—over this distribution. However, this strategy frequently failed because learning a high-dimensional joint distribution is statistically intractable; it requires exponentially large datasets to approximate the complex dependencies between variables.

To overcome this shortcoming, the \gls{acr:ml} community pivoted to \emph{structured learning}, a paradigm that bypasses the estimation of the full distribution. Instead, it trains models to directly minimize a loss defined on the output structure itself, effectively teaching the model to optimize.
These algorithms consist of a statistical model and a solver, or inference engine, integrated into a single differentiable architecture. While the specific combinatorial problems in computer vision—such as finding the most likely pixel grid—differ from \gls{acr:or} challenges like resource allocation, the mathematical foundations are remarkably transferable. For instance, the "smart \gls{acr:pto} plus" (SPO+) loss used in modern \gls{acr:or} is mathematically rooted in the Structured Hinge Loss developed for support vector machines two decades prior. With the advent of deep learning, this paradigm has been revitalized: the linear models of the past have been replaced by deep \glspl{acr:nn}, and the inference step is now framed as a differentiable \gls{acr:co}-layer~\eqref{eq:introSurrogateProblem} within an end-to-end policy.

\paragraph{Alignment with industrial realities.}
In this context, the focus shifts from solving isolated instances to a learning perspective. We assume that instances are drawn from an unknown distribution $\bbP$, and we use a training set to learn a policy $\pi_w$ mapping instances $x$ to solutions $y \in \calY(x)$.
This aligns with industrial practice, where instances often follow a distribution with fixed support, e.g., daily aircraft routing with constant flights but varying loads. Traditional optimization ignores this structure, solving each instance from scratch. A learning-based perspective exploits historical patterns to minimize expected costs across the distribution. Furthermore, this framework often confines combinatorial complexity to a linear optimization step, leveraging efficient linear solvers. This shifts the computational burden to the offline learning phase, leaving a simple linear problem on-the-fly, which is ideal for industrial settings where rapid decisions are crucial.

\subsection{\gls{acr:coaml} within the Decision-Making Landscape}

To position \gls{acr:coaml} relative to other decision paradigms, we use \gls{acr:cso}~\eqref{eq:introCSO} as a common baseline and describe alternatives as policies within this framework.
A key distinction lies in the division between \emph{on-the-fly} computation, i.e., when a new $x$ arrives, and offline computation.
For \gls{acr:coaml} policies, on-the-fly computation involves evaluating the \gls{acr:nn} $\varphi_w(x)$ and solving the \gls{acr:co} layer~\eqref{eq:introSurrogateProblem}. Offline computation, which is far more intensive, involves solving learning problems like~\eqref{eq:IntroEmpiricalCostMinimization} or~\eqref{eq:IntroDecisionFocusedSupervisedLearning}.
Most \gls{acr:cso} policies involve both online \gls{acr:co} solving and offline computation, but differ in key characteritics, see Table~\ref{tab:paradigms-properties}.

\paragraph{Deterministic \gls{acr:co}.}
Here, a deterministic approximation of~\eqref{eq:introCSO} is built and solved on the fly.
This generally involves estimating a single nominal scenario $\bar \xi$, e.g., by using the mean, and solving:
$$ \min_{y \in \calY(x)} c(x,y,\bar \xi).$$
In the terminology of our surrogate framework, the parameters $\theta$ are fixed to $\bar \xi$. While computationally efficient, this approach ignores the dispersion of $\bbP$, often leading to brittle solutions.

\paragraph{Stochastic \gls{acr:co}.}
This approach addresses the flaws of deterministic approximations by directly solving the stochastic optimization problem~\eqref{eq:introHarProblemConditionalExpecation}.
Since $\bbP$ is unknown, a generative model $p(x,\xi)$ or discriminative model $p(\xi|x)$ is typically learned offline.
Given that $c$ is often non-linear or the feasible set complex, the problem is generally intractable. Practitioners therefore resort to \emph{\gls{acr:saa}}:
$$ \min_{y \in \calY(x)} \frac1N\sum_{i=1}^N c(x,y,\xi_i), $$
where $\xi_i$ are i.i.d. samples from $p(\cdot|x)$. However, as the decision space grows, \gls{acr:saa} remains computationally prohibitive for real-time applications.

\paragraph{\Gls{acr:pto}.}
This paradigm decouples learning and optimization into two sequential stages and is closely related to the previous two paradigms. In fact, it can be seen as as a special case of deterministic \gls{acr:co}, where one uses a point forecast to create the nominal scenario, allowing to better capture context and uncertainty dynamics.
Specifically, a predictive model $\varphi_w$ is trained to minimize a decision-agnostic loss, e.g., an \gls{acr:mse}, against the ground truth $\xi$, purely focusing on prediction accuracy in the first stage.
In the second stage, the predicted value $\hat \xi = \varphi_w(x)$ is plugged into the deterministic optimization problem $\min_{y \in \calY(x)} c(x, y, \hat \xi)$.
While this retains the computational efficiency of deterministic approaches, it suffers from a fundamental \emph{loss-cost mismatch}: 
\begin{table}[!hb]
	\footnotesize
	\centering
	\caption{Comparative analysis of decision-making paradigms.}
	\label{tab:paradigms-properties}
	\begin{tabular}{p{2.2cm}p{2.8cm}p{3.6cm}ccl}
		\toprule
		\textbf{Paradigm} & \textbf{Online \mbox{Computation}} & \textbf{Primary \mbox{Constraint}} & \textbf{Coincidence} & \textbf{Learning} & \textbf{Loss Target} \\
		\midrule
		\textbf{Deterministic \gls{acr:co}} & Solve nominal instance & Ignores uncertainty & Yes & None & N/A \\
		\addlinespace[0.3em]
		\textbf{Stochastic \gls{acr:co}} & Solve \gls{acr:saa} of stochastic problem & Intractable for large $\calY(x)$ & Yes & Gen. & N/A \\
		\addlinespace[0.3em]
		\textbf{\gls{acr:pto}} & Solve predicted instance & Loss-cost mismatch & No & DA & Prediction Error \\
		\addlinespace[0.3em]
		\textbf{Smart-\gls{acr:pto}} & Solve predicted instance & Restricted to linear objectives & Yes$^\dagger$ & DF & Regret / Cost \\
		\addlinespace[0.3em]
		\textbf{Structured Prediction} & MAP Inference & Limited to tractable structures & No & DF & Imitation / Cost \\
		\addlinespace[0.3em]
		\textbf{\gls{acr:coaml}} & Solve learned surrogate & Non-convex / Hard training & No & DF & Imitation / Cost \\
		\bottomrule
		\multicolumn{6}{p{0.97\textwidth}}{\scriptsize Abbreviations: DA--Decision-Agnostic Learning; DF--Decision-Focused Learning; Gen.--Generative Learning; MAP--Maximum a Posteriori; Coincidence: Denotes if the online surrogate problem is mathematically equivalent to the underlying stochastic objective (or its \gls{acr:saa}). $^\dagger$ Coincidence holds strictly only for linear objectives where $\bbE[c(x,y,\xi)] = c(x,y,\bbE[\xi])$.}
	\end{tabular}
\end{table}
the training objective ignores the downstream optimization landscape; accordingly, small prediction errors can result in arbitrarily large decision costs.

\paragraph{Smart \gls{acr:pto}.}
When the objective is linear in the uncertainty, the deterministic approximation (using the conditional expectation) and the stochastic problem coincide mathematically:
$$\min_{y\in \calY(x)}\bbE\big[c(x,y,\bfxi)\big|x\big] = \min_{y\in \calY(x)}\underbrace{\bbE\big[\bfxi|x\big]}_{\bar \xi}^\top y = \min_{y\in \calY(x)}c(x,y,\bar \xi).$$
The smart \gls{acr:pto} literature exploits this property, training a \gls{acr:nn} $\varphi_w$ to predict $\theta = \bar \xi = \varphi_w(x)$. Unlike standard \gls{acr:pto}, it uses a decision-focused loss such as SPO+ rather than a prediction loss, cf. \citet{elmachtoub2022smart}. This ensures that even if the prediction $\varphi_w(x)$ is imperfect, it induces a near-optimal decision $y$.

\paragraph{Structured prediction.}
Structured learning applications from the early 2000s share the mathematical form of \gls{acr:cso} but differ in intent.
Often, the true cost $c(x,y,\xi)$ is unknown or hard to specify, 
e.g., the "cost" of a wrong pixel in image segmentation, but expert demonstrations are available. The surrogate problem~\eqref{eq:introSurrogateProblem} is typically a maximum a posteriori (MAP) inference problem over a graphical model. While these methods pioneered decision-focused learning when following an \gls{acr:il} setting, the underlying combinatorial structures were often simpler than those found in industrial~\gls{acr:or}.

\paragraph{Neural \gls{acr:co}.}
Neural \gls{acr:co} has gained significant traction as an alternative paradigm. In the \gls{acr:cso} context, its policies function as constructive heuristics rather than solvers.
For example, in the \gls{acr:tsp}, a sequence-to-sequence network predicts the next city to visit given a partial tour.
The surrogate problem is implicit: there is no explicit optimization instance solved on the fly, but rather the execution of a parameterized policy trained via \gls{acr:rl}.
We exclude Neural \gls{acr:co} from this review as it discards the exact solvers that \gls{acr:coaml} aims to leverage.

\paragraph{Synthesis: the \gls{acr:coaml} paradigm.}
\gls{acr:coaml} emerges as a generalization of these approaches, designed to handle the complexity of \gls{acr:or} problems where Smart \gls{acr:pto}'s linearity assumptions do not hold and Stochastic \gls{acr:co}'s computational costs are prohibitive.
By treating the solver as a parameterized surrogate layer $\max \kappa(\theta, y)$, \gls{acr:coaml} decouples the learned parameters $\theta$ from the physical predictions $\bar \xi$. The parameters $\theta$ do not need to represent expected demands or costs; rather, they are latent "knobs" tuned by the network to force the solver into a low-cost configuration. This allows \gls{acr:coaml} to utilize the speed of deterministic solvers while accounting for complex stochastic interactions through decision-focused training, effectively closing the loop between data and decision.

\subsection{Scope and Contributions}

Recent literature has extensively charted the intersection of \gls{acr:ml} and \gls{acr:or}. Comprehensive surveys, such as those by \cite{sadana2024survey} and \cite{mandiDecisionFocusedLearningFoundations2024}, map the broad landscape of data-driven optimization, covering a vast array of modeling approaches and theoretical guarantees. Complementing these, tutorials by \cite{misicDataAnalyticsOperations2020} and \cite{kotary2021} provide technical deep dives into specific methods for accelerating constrained optimization.
In contrast, this paper adopts a distinct "tutorial-survey" perspective. Rather than attempting to cover the entire spectrum of contextual decision-making, we provide a focused, comprehensive primer on a specific, high-impact paradigm: \emph{Combinatorial Optimization Augmented Machine Learning (COAML)}.

The \gls{acr:coaml} paradigm fundamentally rethinks the interface between prediction and optimization. Although we introduced \gls{acr:coaml} through the theoretical lens of \gls{acr:cso}, its applicability extends far beyond this setting. The core innovation—embedding combinatorial oracles as differentiable layers within learning architectures—allows for end-to-end training in diverse environments. This ranges from learning surrogate costs for static NP-hard problems to enabling \gls{acr:srl} policies in multi-stage environments. Accordingly, this paper is designed to serve as both a tutorial for newcomers and a structured reference for experts. Our contributions are organized as follows:
\begin{description}
    \item[Formalization:] We rigorously define \gls{acr:coaml} architectures  and distinguish them from related \gls{acr:pto} and neural \gls{acr:co} approaches.
    \item[Applications:] We illustrate the paradigm's versatility through key application domains, demonstrating its value in both industrial operations and scientific computing.
    \item[Methodology:] We outline the essential building blocks, with a focus on the techniques required to differentiate through discrete optimization layers.
    \item[Taxonomy:] We review existing work and best practices, organizing the literature into a structured taxonomy that categorizes methods by their learning signal and solver integration.
    \item[Guidelines:] Based on the taxonomy, we provide guidelines on how to leverage the available tools to design succesful \gls{acr:coaml} approaches. 
    \item[Open challenges:] We identify critical gaps in theory and practice, proposing directions for future research.
\end{description}

Beyond our primary scope, we acknowledge a related line of research focusing on \emph{continuous} optimization layers. While this shares challenges with \gls{acr:coaml}, such as gradient estimation and implicit differentiation, they require distinct mathematical techniques. We briefly outline their relationship to \gls{acr:coaml} in the conclusion to provide a complete picture.  Finally, we note that this paper makes use of various acronyms. To aid readability, we provide an overview of these acronyms in the appendix.
\section{Formalization \& Problem Classes}\label{sec:formalizationApplication}

As introduced earlier, \gls{acr:coaml} architectures are \glspl{acr:nn} that terminate with a \gls{acr:co}-layer. These architectures define policies $\pi_w$ that map a context $x \in \calX$ to a decision $y \in \calY(x)$, or potentially a conditional distribution over $\calY(x)$, parameterized by weights $w$.

While we initially presented these architectures in the context of \gls{acr:cso}, where the goal is to minimize an expected cost $\bbE[c(x, y, \boldsymbol{\xi})]$ given a context realization $x$ and random uncertainty $\boldsymbol{\xi}$, they are in fact much more general. They provide a flexible mechanism for parameterizing policies over combinatorial spaces that can be learned from data.
This versatility allows \gls{acr:coaml} to be applied to a wide range of problem classes, ranging from hard deterministic problems to multi-stage stochastic optimization.

In the following, we give an overview of problem classes where the \gls{acr:coaml} framework can be useful to derive efficient algorithms.
These problem classes will serve as running examples throughout the paper, and we will return to them in Section~\ref{sec:stateoftheart} when mapping algorithms and applications to our taxonomy of \gls{acr:coaml} approaches.

\subsection{Learning heuristics for hard \gls{acr:co} problems.}
\label{sub:heuristicsForHardProblems}
Industries today have access to vast amounts of data, which they leverage to enhance the performance and resilience of their operational processes. Achieving this potential requires algorithms that not only optimize these processes but also scale effectively. The primary improvement potential of optimization algorithms stems from their ability to reduce marginal costs. Consider, for instance, a routing application. Even the most advanced routing algorithm incurs high per-request costs if only three delivery requests exist, as the vehicle must still complete a tour starting and returning to a depot. In contrast, with one thousand delivery requests, the vehicle can consolidate deliveries within the same neighborhood, significantly lowering the marginal cost per request. However, to realize these gains, the underlying algorithm must scale efficiently to handle such large instances.

The \gls{acr:co} community has therefore devoted substantial efforts to designing algorithms that scale well for large instances, often accepting simplified objective functions to maintain tractability. As a result, problems with linear objectives are pervasive in \gls{acr:co}. Decades of research have yielded practically efficient algorithms for a wide range of applications. However, these methods frequently become intractable when the objective function incorporates complex phenomena or resilience considerations, which introduce nonlinearities or stochastic elements. In such contexts, we encounter a fundamental discrepancy: while the problem requires scalable algorithms for complex combinatorial structures, existing methods typically only achieve true scalability for simpler linear approximations.

This setting can be formally cast as a \gls{acr:cso} problem. Here, the context $x$ represents the instance of the hard problem, e.g., the set of requests. 
In many cases, the difficult problem is stochastic, and the uncertainty $\bfxi$ represents the complex or stochastic elements, e.g., delays or disruptions, that make the true cost $\bbE_{\bfxi}[c(x, y, \bfxi)]$ difficult to optimize directly. If the problem is deterministic, then $\xi$ can be skipped. The \gls{acr:coaml} policy uses a simpler, scalable combinatorial layer, e.g., a deterministic linear solver, to generate solutions $y$, while the learning process adjusts the parameters of this layer to minimize the complex true cost $\bbE_{\bfxi}[c(x, y, \bfxi)]$.

Complexity theory indicates that we cannot expect a policy $\pi_w$ to perform well for every possible instance $x$. Nevertheless, practical instances often exhibit specific structures captured by the distribution $\bbP$ of $x$, and possibly $\bfxi$. In many applications, this structure allows us to achieve excellent performance in practice, which translates into a low risk, i.e., effectively solving~\eqref{eq:IntroEmpiricalCostMinimization}.

\begin{example}\label{ex:stochastic_vsp} \textbf{Stochastic vehicle scheduling.}
    The \gls{acr:vsp} is a classical \gls{acr:co} problem in which a fleet of vehicles must serve a set of timed requests. The objective is to construct sequences of requests for each vehicle that collectively cover all requests at minimal cost. A common application of the \gls{acr:vsp} arises in aircraft routing, where airlines must assign aircraft to scheduled flights.

    When the objective function is simply the sum of the arc costs, the \gls{acr:vsp} can be formulated and solved efficiently as a minimum cost flow problem. In this formulation, the graph’s vertices correspond to requests, and the arcs $a \in A$ represent pairs of requests that can be consecutively served by a vehicle within a feasible route. The rich literature and efficient algorithms for minimum cost flow problems make this deterministic formulation well-suited for practical applications.

    However, practitioners often encounter stochastic variants of the \gls{acr:vsp}. For example, delays in flights can propagate to subsequent legs of an aircraft’s route. Airlines therefore aim to find aircraft routing solutions that minimize the \emph{expected} total delay cost, taking into account the randomness inherent in operational disruptions. These stochastic variants are considerably more challenging from a computational standpoint, and there is generally no algorithm that scales well enough for industrial applications in these settings.

    To address this challenge, \citet{parmentierLearningApproximateIndustrial2021} proposed a \gls{acr:coaml}-based approach to approximate the \gls{acr:svsp} by reparametrizing the deterministic \gls{acr:vsp}. As illustrated in Figure~\ref{fig:stovsp_pipeline}, a statistical model $\varphi_\vw$ predicts linear arc costs $\theta = (\theta_a)_a$ that are tailored to yield good solutions for the stochastic variant of the problem. A deterministic \gls{acr:vsp} solver then computes a solution $y$ using these learned arc costs.

    This approach has a significant advantage in industrial practice: it leverages the existing deterministic solver, including all the specialized industrial constraints already incorporated into it. By learning to adjust only the arc costs $\theta$ while keeping the solver unchanged, 
    \gls{acr:coaml} enables practitioners to integrate data-driven adaptations to uncertainty without having to overhaul their existing optimization infrastructure.
     \begin{figure*}[!htb]
    	\includegraphics[width=\textwidth]{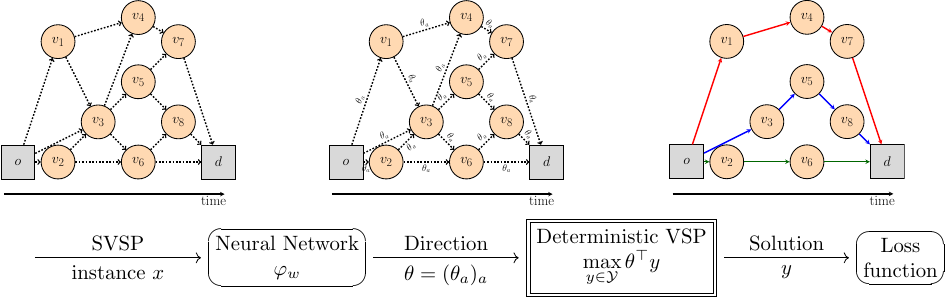}
    	\caption{Neural network with a \gls{acr:co}-layer for the stochastic vehicle scheduling problem.
    		Vertices represent requests. Dotted arrows represent arcs $a$, which are pairs of requests that can be operated in a sequence. The colored paths give vehicle routes in the solution returned. Image adapted from~\citet{dalle2022learning}.}
    	\label{fig:stovsp_pipeline}
    	\vspace{-6mm}
    \end{figure*}
\end{example}

\subsection{Structured Learning}

The roots of modern \gls{acr:coaml} can be traced back to the \emph{structured learning} community. In the early 2010s, before the widespread adoption of end-to-end deep learning, \gls{acr:co}-layers became the state of the art for tasks involving complex output dependencies, such as image segmentation or natural language parsing~\citep[cf.][]{nowozinStructuredLearningPrediction2011}. Unlike standard classification, where predictions are treated as independent scalars, these tasks require predicting a configuration of variables that must respect structural constraints—effectively solving a combinatorial problem to ensure the output is valid and coherent.

This problem class can be seamlessly viewed through the lens of \gls{acr:cso}, albeit with a specific interpretation of the variables. In this case, the context $x$ represents the raw input, e.g., an image or sentence, while the ``uncertainty'' $\bfxi$ degenerates to the unknown ground truth structured output $\bar y$, e.g., the correct segmentation map. The cost function is effectively a supervised loss function $c(x, y, \bfxi) = \calL(y, \bar y)$ that measures the discrepancy between the predicted structure $y$ and the target $\bar y$. Consequently, this setup corresponds to the \gls{acr:sl} setting defined earlier, where we observe pairs $(x, \bar y)$ and aim to minimize the expected loss over the data distribution.

Beyond its historical significance, this field has provided the algorithmic bedrock for \gls{acr:coaml}. Many of the differentiation techniques used in \gls{acr:or} today, e.g., perturbation-based differentiation and \gls{acr:fy}-losses, originated here~\citep{blondelLearningFenchelYoungLosses2020, berthetLearningDifferentiablePerturbed2020}. The community remains vibrant, with applications expanding into domains like information retrieval, where ranking problems are modeled as optimization over the permutahedron. For an up-to-date and comprehensive introduction to these foundational methods, we refer to \citet{blondel_edpbook}.

\subsection{Contextual Stochastic Optimization}

Although the introductory section established the basic mechanics of \gls{acr:cso}, its full scope warrants further elaboration to appreciate its versatility. In this framework, the context $x$ plays a pivotal, dual role.
First, it defines the structural parameters of the optimization instance itself, determining the feasible set $\calY(x)$. This dependence allows the framework to accommodate dynamic environments where the constraints change from instance to instance, a standard requirement in operational problems.
Second, the context $x$ serves as a carrier of predictive information regarding the uncertainty~$\bfxi$.
Unlike traditional stochastic optimization approaches, which often struggle to condition decisions on high-dimensional or complex feature spaces, \gls{acr:coaml} seamlessly integrates this contextual information into the decision pipeline. By leveraging the representational power of \glspl{acr:nn}, the policy can exploit subtle correlations between the context and the uncertainty to minimize expected costs.

\begin{continuance}{ex:stochastic_vsp}
    Returning to the \gls{acr:svsp} of Example~\ref{ex:stochastic_vsp}, consider that the context $\tilde x$ comprises the weather forecast, a signal highly correlated with the flight delays $\bfxi$. Furthermore, the set of flights to be covered—and consequently the feasible solution set $\calY(x)$—varies naturally from week to week, making the optimization domain itself dependent on the instance $x$.
\end{continuance}

\subsection{Two-Stage Stochastic Optimization}

The \gls{acr:coaml} framework readily extends to \emph{two-stage stochastic optimization}, a setting where decisions are partitioned into immediate ``here-and-now'' actions and subsequent ``wait-and-see'' recourse actions.
This problem class can be modeled as:
\begin{equation*}\label{eq:two_stage_stochastic_optimization}
    \min_{y \in \calY}  \bbE\Big(\tilde c(x, y, \bfxi)\Big) \quad \text{where} \quad \tilde c(x, y, \bfxi) = \argmin_{z \in \calZ(x,y,\bfxi)} c(x,y,z,\bfxi). 
\end{equation*}
In this formulation, the cost function $\tilde c$ represents the value of the optimal second-stage recourse $z$, which is determined after the uncertainty $\bfxi$ is realized.
The variable $x$ acts as the context, encoding both the instance definition and all information available for the first-stage decision.
Structurally, these problems are a specific instance of the general \gls{acr:cso} framework where the immediate cost is augmented by the outcome of a downstream optimization problem.

Applying \gls{acr:coaml} in this setting typically involves learning a policy for the first stage that implicitly anticipates the expected recourse.
Consequently, the combinatorial layer often involves solving a linear problem solely on the first-stage decision, effectively compressing the two-stage complexity into a single learned representation~\citep[cf.][]{dalle2022learning}. Alternatively, if the second-stage feasible set $\calZ(x,y,\bfxi)$ is independent of the uncertainty $\bfxi$, the layer can be constructed as a linear problem over both decision stages~\citep{parmentierLearningStructuredApproximations2021}.  

\subsection{Data-Driven Optimization from Raw Inputs}

A particularly powerful application of \gls{acr:coaml} lies in \emph{data-driven optimization} where the parameters of the optimization problem must be inferred from high-dimensional, unstructured data.
In this setting, the context $x$ typically represents raw observations, such as satellite imagery, textual descriptions, or complex sensor streams.
The uncertainty $\bfxi$ corresponds to the underlying ground-truth parameters of the optimization problem, such as edge costs, capacities, or utilities, which are not directly observable.
The cost function $c(x, y, \bfxi)$ evaluates the objective value of the solution $y$ under these true parameters.

Traditional \gls{acr:or} architectures require a manual feature engineering step to convert raw data into structured coefficients.
\gls{acr:coaml}, by contrast, leverages deep \glspl{acr:nn} $\varphi_w$ to learn a direct mapping from raw perception to combinatorial reasoning.
The network acts as a predictive model that extracts the latent parameters $\theta \approx \bfxi$ from the raw context $x$, which are then immediately processed by the combinatorial layer.
This end-to-end approach ensures that the feature extraction is optimized specifically for the downstream decision quality, rather than for generic reconstruction accuracy.

\begin{example}\label{ex:paths_in_images}\textbf{Perception-based Shortest Paths.}
    \citet{vlastelicaDifferentiationBlackboxCombinatorial2020} introduce the \emph{Warcraft Shortest Path} benchmark, a task that has become a standard reference for evaluating differentiable solvers. The problem involves computing shortest paths across terrain maps represented solely as pixel images.
    Here, the context $x$ is the raw image of the map, and the neural network $\varphi_w$, typically a convolutional architecture, must visually infer the movement cost $\bfxi$ associated with each terrain type (e.g., water, forest, grass).
    The oracle $\hat y$ corresponds to Dijkstra's algorithm, which takes these predicted costs and outputs the optimal path.
    Crucially, the supervision signal is often defined on the final path rather than the individual tile costs, forcing the network to learn the physics of the terrain purely through the optimization loop.
\end{example}

\subsection{Multistage Stochastic Optimization}\label{sec:multistageStoOpt}
Let us conclude this section with a flagship application of \gls{acr:coaml} in \gls{acr:or}: multistage stochastic optimization problems in a combinatorial setting. Unlike the previous examples, this setting is best described as a \gls{acr:mdp} rather than a static \gls{acr:cso} problem. The system evolves in discrete time steps $t=0,1,\ldots,T$. At each time step $t$, the system resides in a state $x_t \in \calX_t$, and the decision maker must select a decision $y_t \in \calY_t$. The system then transitions to a new state $x_{t+1}$ according to a (possibly unknown) transition probability $p(x_{t+1} \mid x_t, y_t)$. The decision maker incurs a cost $c_t(x_t, y_t)$ that depends on the current state and the chosen decision. A solution to this problem is a policy $\pi : x_t \in \calX_t \longmapsto y_t \in \calY_t$ that minimizes the expected cumulative cost:
\begin{equation}\label{eq:multistage_stochastic_optimization}
    \min_{\pi} \bbE_{\pi}\Big[\sum_{t=0}^T c_t(\bfx_t, \bfy_t)\Big| \pi\Big].
\end{equation}
Analogously to previous settings, policies do not need to be deterministic, and can be turned into conditional probability distributions.

In this setting, classical dynamic programming provides exact solutions when the state space $\calX_t$ is sufficiently small to permit enumeration and the optimization over $\calY_t$ remains tractable.
Outside of this regime, existing methodologies typically compromise on one dimension of complexity.
The multi-stage stochastic optimization literature offers robust methods for handling large, complex action spaces $\calY$, e.g., via stochastic dual-dynamic programming, but generally assumes the state space $\calX$ has moderate dimension.
Conversely, deep \gls{acr:rl} excels at processing high-dimensional state spaces $\calX$, e.g., images, but often struggles when the action space $\calY$ is large, discrete or combinatorial.
A significant gap therefore remains for problem settings where both $\calX$ and $\calY$ are large and structured.
\gls{acr:coaml} architectures naturally address this dual complexity by assigning the burden of state representation to the neural network and the burden of action feasibility to the combinatorial layer.

\begin{example}\textbf{Dynamic vehicle routing}\label{ex:dynamic_vehicle_routing}
    The 2022 EURO meets NeurIPS vehicle routing competition~\citep{euromeetsneurips2022} centered on a \gls{acr:dvrp}. The task involved a rapid delivery service using capacitated vehicles to serve customer requests originating from a depot. Each request had to be served within a specified time window. Requests arrived dynamically, and vehicles were dispatched in waves to serve them. At each wave's decision time $t$, the system state $x_t$ consists of the set of requests that have not yet been served. The decision $y_t$ involves selecting the subset of requests to be served by the vehicles dispatched at time $t$, as well as the corresponding routing plan.

    The objective is to find a policy $\pi$ that minimizes the expected total routing cost. \citet{batyCombinatorialOptimizationEnrichedMachine2024} observe that the feasible set $\calY_t(x_t)$ at each decision epoch $t$ can be formulated as a prize-collecting \gls{acr:cvrp}. In this variant of the \gls{acr:cvrp}, serving each request $v$ yields a prize $\theta_v$, and the goal is to maximize the total profit, defined as the sum of the collected prizes minus the routing cost. However, in the \gls{acr:dvrp} considered here, there is no explicit notion of prizes—only routing costs. To address this, the authors construct a policy $\pi_w$ as illustrated in Figure~\ref{fig:dynamic_vrp}, in which a neural network $\varphi_w$ predicts the prize values $(\theta_v)_v$. The prize-collecting \gls{acr:cvrp} is then solved using these predicted prizes to produce the routing solution $y_t$.

    This approach demonstrated a significant performance improvement and ultimately won the competition by a considerable margin.
\end{example}

\section{Methodology}
Having introduced the formal setting and problem classes, we now turn to the methodological 
building blocks of \gls{acr:coaml} pipelines: architectures, and learning algorithms.

Crucially, the training methodology is not dictated by the problem class, but by the nature of the learning signal available during the offline training phase. We distinguish three primary learning paradigms:
\begin{description}
    \item[Supervised learning:] When a dataset of expert decisions or ground-truth optima $(x_i, \bar y_i)$ is available, the policy can be trained to replicate these targets by minimizing a divergence measure or loss $\calL(\pi_w(x_i), \bar y_i)$, a process commonly referred to as \gls{acr:il}.
    \item[Empirical cost minimization:] When explicit targets are unavailable but the decision maker has access to the cost function $c$ and historical realizations $\xi_i$ of the uncertainty $\bfxi$, the policy can be trained to directly minimize the empirical cost $\sum c(x_i, \pi_w(x_i), \xi_i)$. This approach is particularly suitable for white-box settings where the system physics are known.
    \begin{figure*}[!htb]
    	\includegraphics[width=\textwidth]{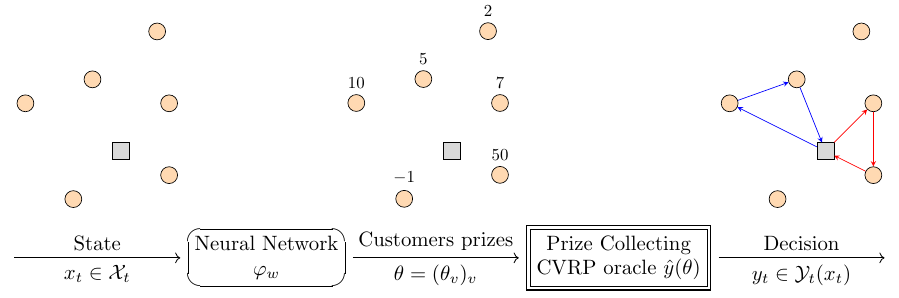}
    	\caption{\gls{acr:coaml} policy for the EURO-NeurIPS \gls{acr:dvrp}. Yellow nodes represent request, the grey square the depot, arrows the dispatched routes. The illustration is a courtesy of Léo Baty.}
    	\label{fig:dynamic_vrp}
    \end{figure*}
    \item[Reinforcement learning:] When neither targets nor an explicit cost function are available and the cost can only be observed as a scalar reward signal following interaction with an environment, the policy can be trained using \gls{acr:rl} techniques. This is the standard approach for sequential or black-box environments.
\end{description} 
When considering multistage stochastic optimization problems, these learning paradigms adapt to the sequential nature of the problem:
\begin{description}
    \item[Imitation learning:] If expert trajectories or optimal policy demonstrations are available, the policy can be trained to clone this behavior using \gls{acr:sl}, effectively reducing the sequential problem to a series of static classification or regression tasks.
    \item[Reinforcement learning:] When the transition dynamics are unknown, complex, or black-box, the policy must be trained via interaction. The agent observes the state $x_t$, generates a combinatorial action $y_t$, and updates parameters to minimize cumulative costs based on the sparse reward signal.
    \item[Direct policy search:] This is the dynamic equivalent of \gls{acr:ecm}. Given a differentiable simulator or a fixed dataset of historical trajectories, we can optimize the policy parameters to directly minimize the total empirical cost accumulated over the horizon.
\end{description}

The rest of the section introduce the architectures and the learning paradigms.
Among these, \gls{acr:ecm} plays a particularly 
central role in \gls{acr:coaml}, serving as both the conceptual foundation and the source of many 
open challenges. We therefore devote more space to \gls{acr:ecm} than to the other components.
\subsection{Architectures}
\label{sec:architectures}
\begin{figure}[!hb]
	\centering
	\includegraphics[width=0.9\textwidth]{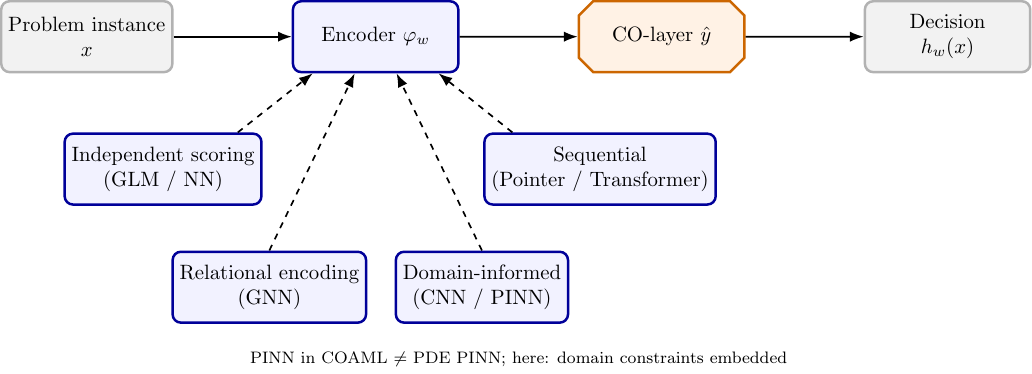}
	\caption{Schematic \gls{acr:coaml} pipeline. An input instance $x$ is encoded by a statistical model $\varphi_w$,
		which parametrizes a \gls{acr:co} oracle $\hat y$ to produce a decision $\pi_w(x)$. 
		Different encoder architectures can be used depending on the problem class: independent scoring models, 
		\glspl{acr:gnn}, sequential/constructive models, or domain-informed encoders.}
	\label{fig:coaml_pipeline}
\end{figure}

When applying \gls{acr:coaml} to a specific problem, the first task is to design the architecture of the policy~$\pi_w$. 
This implies choosing the \gls{acr:co}-oracle $\hat y$ and the statistical model $\varphi_w$. 
The choice of $\hat y$ is application dependent and dictated by the structure of the feasible set $\calY(x)$ as well as the algorithms that scale well for this class of problems. 
In contrast, the choice of $\varphi_w$ is more generic: it determines how contextual information and problem features are encoded into parameters $\theta$ that interact with the combinatorial oracle, see Figure~\ref{fig:coaml_pipeline}. 
The design challenge is therefore not only predictive accuracy but the alignment of the architecture with the downstream \gls{acr:co}-layer.

\paragraph{Independent scoring models.}
A first family of approaches relies on assigning independent scores to the dimensions of the vector space where $\calY(x)$ lies.
Each dimension $i$ is represented by a feature vector $\phi(i,x)$, and a simple parametric model assigns it a score $\theta_i$. 
A \gls{acr:glm} is the simplest such choice:
\begin{equation}\label{eq:parallel_generalized_linear_model}
    \varphi_w(x)  = (\theta_i)_{i \in \{1,\ldots,d(x)\}} \quad \text{where} \quad \theta_i = w^\top \phi(i,x).
\end{equation}

These scores parametrize the \gls{acr:co}-layer, which then computes a feasible solution by optimizing over~$\calY(x)$. 
This approach is highly scalable and interpretable, but it ignores structural dependencies between elements. 

\begin{continuance}{ex:stochastic_vsp}
    In the \gls{acr:svsp}, the dimensions of the flow polytope correspond to the arcs $a$ of the digraph. 
    A vector of features $\phi(a,x)$ summarizes the characteristics of each prospective connection $a$ in the instance~$x$, and a \gls{acr:glm} predicts arc costs as $\theta_a = w^\top \phi(a,x)$ \citep{parmentierLearningApproximateIndustrial2021}. 
    The deterministic \gls{acr:vsp} solver then uses these predicted arc costs to generate routing solutions.
\end{continuance}

A natural extension of this idea is to replace the linear model by a feed-forward \gls{acr:nn} $\psi_w$, applied in parallel to each dimension:
\begin{equation}\label{eq:parallel_neural_network}
    \varphi_w(x)  = (\theta_i)_{i \in \{1,\ldots,d(x)\}} \quad \text{where} \quad \theta_i = \psi_w \circ \phi(i,x).
\end{equation}
This increases expressive power while retaining scalability, and is often used in \gls{acr:cso} tasks.

\paragraph{Graph neural networks.}
Many combinatorial problems exhibit strong relational structure. 
\glspl{acr:gnn} exploit this by encoding elements as vertices of a graph $G$, with edges capturing feasible interactions such as temporal or spatial adjacency. 
Message passing aggregates local information into context-aware embeddings, which are then mapped to scores $\theta_i$: 
\begin{equation}\label{eq:graph_neural_network} 
    \varphi_w(x)  = \psi_w (\tilde\phi) \quad \text{where} \quad \tilde\phi = \big(\phi(i,x)\big)_{i \in \{1,\ldots,d(x)\}}. 
\end{equation}

\begin{continuance}{ex:dynamic_vehicle_routing}
    \citet{batyCombinatorialOptimizationEnrichedMachine2024} won the EURO--NeurIPS vehicle routing challenge using a sparse \gls{acr:gnn} encoder. 
    Requests $v$ were represented as nodes of the graph $G$, with edges linking geographically and temporally close requests. 
    The predicted scores $\theta_v$ then parameterized a prize-collecting \gls{acr:vrp} oracle, yielding high-quality routing solutions.
\end{continuance}

\paragraph{Sequential and autoregressive architectures.}
Some combinatorial problems, such as routing or scheduling, are inherently sequential. 
Pointer networks and Transformer-based models capture such order dependencies by conditioning each prediction on the partial solution constructed so far. 
These autoregressive architectures parameterize constructive heuristics or subproblems by producing context-dependent scores $\theta$. 
They have shown strong performance in routing and packing domains, especially when being combined with \gls{acr:il} or \gls{acr:rl} to stabilize training and improve generalization.

\paragraph{Domain-informed encoders.}
In many applications, raw data already encode structural constraints. 
Domain-informed encoders exploit this by embedding prior knowledge into the statistical model. 
Examples include \glspl{acr:cnn} for spatial data, recurrent networks for temporal processes, and \glspl{acr:pinn}. 
We emphasize that in the \gls{acr:coaml} context, ``\gls{acr:pinn}'' does not denote the PDE-oriented \glspl{acr:pinn} common in applied mathematics. 
Rather, it refers to neural architectures that incorporate structural or physical constraints, such as conservation laws or inventory balance relations, directly into the encoder. 
This improves sample efficiency and ensures predictions align with the feasible space of the \gls{acr:co}-layer.

\begin{continuance}{ex:paths_in_images}
    In the path-in-images benchmark introduced by \citet{vlastelicaDifferentiationBlackboxCombinatorial2020}, shortest paths are computed on maps represented as images. 
    A \gls{acr:cnn} $\varphi_w$ predicts arc costs from pixel-based features, and Dijkstra's algorithm serves as the \gls{acr:co}-layer to compute the optimal path. 
\end{continuance}

\begin{example}\label{ex:inventory_vrp}\textbf{Dynamic Inventory Vehicle Routing}
    In the \gls{acr:dirp}, \citet{greif2024combinatorial} employed a domain-informed neural network architecture, classified as a \gls{acr:pinn}, to encode the dynamics of inventory flows and demand evolution. 
    These predictions were used to parametrize a prize-collecting \gls{acr:tsp} oracle. 
    Unlike classical PDE-based \glspl{acr:pinn}, the model here embedded inventory balance constraints directly into the network architecture to better align predictions with feasible routing solutions. 
\end{example}

\subsection{Supervised Learning}
\label{sec:supervised_learning}
We now turn to the \gls{acr:sl} problem, see Equation~\eqref{eq:IntroDecisionFocusedSupervisedLearning}. We consider having a training set of instances $x_1, \ldots, x_n$ with target solutions $\bar y_1, \ldots, \bar y_n$, and consider the following learning problem.
$$ \min_{w}\sum_{i=1}^n \calL\big(\hat y \circ \varphi_w(x_i),\bar y_i\big) $$
Our main task is to design a loss $\calL(y,\bar y)$ on the solution space.
Since $\varphi_w$ is a neural network, we aim for a loss that can be optimized using stochastic gradient descent.

In this section, we fix an $x$ and omit the dependence on $x$ for notational simplicity. We denote $\calY(x)$ by $\calY$, and the dimension of $\theta$ by $d$.
Let $\calC$ be the polytope $\conv(\calY)$. 
Remark that since $\hat y$ does not depend on $w$, fixing $\theta$ amounts to fixing $y$. 
Most approaches therefore define a loss $\ell$ between the direction $\theta$ and the target solution $\bar y$, leading to the learning problem
\begin{equation}\label{eq:supervised_learning_primal_dual_loss}
    \min_w \sum_{i=1}^{n}\ell\big(\varphi_w(x_i),\bar y_i\big), 
\end{equation}
where $\ell(\theta,\bar y) = \calL(\hat y(\theta),\bar y)$.
We illustrate the main difficulty when defining such a loss in Figure~\ref{fig:piecewise_constant}.
Consider a vertex $y$ of $\calC$. The output of $\hat y$ is $y$ for any $\theta$ in the interior of the normal cone of $y$.
As a consequence, $\hat y$ is piecewise constant on each cone of the \emph{normal fan} of $\calC$, which is the partition of $\bbR^{d}$ into normal cones for the vertices of $\calC$. In the structured learning literature, authors often replace this piecewise constant loss by the structured Hinge loss \citep{nowozinStructuredLearningPrediction2011}[Chapter 6], a surrogate upper bound that is convex in $\theta$. It has recently enjoyed a new popularity as the SPO+ loss~\citep{elmachtoubSmartPredictThen2021}.
In the following, we go beyond structured hinge losses and present a recent alternative that enjoys nice geometric properties and is convenient for practical applications.

\paragraph{Regularized prediction, Fenchel-Young loss, and Bregman divergence.}

Since maximizing a linear objective on the vertices $\calY$ of a polytope $\calC = \conv(\calY)$ is equivalent to maximizing it on $\calC$, we can directly solve:
\begin{equation}\label{eq:combinatorial_optimization_oracle_polytope}
    \hat y(\theta) = \argmax_{y \in \calC} \theta^\top y.
\end{equation}
The non-differentiability of $\hat y$ comes from the fact that the optimal solution may ``jump'' from one vertex to another when $\theta$ crosses the boundary between two normal cones of the normal fan. This behavior disappears when we regularize~\eqref{eq:combinatorial_optimization_oracle_polytope} with a smooth and strictly convex function~$\Omega$, which pushes the optimal solution to the interior of the polytope:
\begin{equation}\label{eq:regularized_combinatorial_optimization_oracle_polytope}
     \argmax_{y \in \calC} \theta^\top y - \Omega(y).
\end{equation}
When defining a loss $\ell$ between $\theta$ and $\bar y$, we would like $\ell(\theta,\bar y)$ to be small when $\bar y$ is ``close to'' being an optimal solution of~\eqref{eq:regularized_combinatorial_optimization_oracle_polytope} for $\theta$.
It is then natural to define the \emph{\gls{acr:fy} loss} as the non-optimality of $\bar y$ when being a solution to~\eqref{eq:regularized_combinatorial_optimization_oracle_polytope} for $\theta$:
\begin{figure}[!hb]
	\centering
	\begin{tabular}{rcl}
		\begin{tikzpicture}[scale=1.9]
			
			\coordinate (A) at (0.3,0);
			\coordinate (B) at (1,0);
			\coordinate (C) at (1.2,0.6);
			\coordinate (D) at (0.8,1);
			\coordinate (E) at (0.2,1);
			\coordinate (F) at (-0.2,0.6);
			
			\draw (A) -- (B) -- (C) -- (D) -- (E) -- (F) -- cycle;
			
			\node[below left] at (A) {$y_5$};
			\node[below right] at (B) {$y_6$};
			\node[above] at (C) {$y_1$};
			\node[above right] at (D) {$y_2$};
			\node[above left] at (E) {$y_3$};
			\node[above left] at (F) {$y_4$};
			
			\coordinate (Center) at (C);
			
			\draw[dashed, fill=red!5, draw=red!60!black] (Center) -- ($(Center)!0.75cm!95:(B)$) arc (0:45:1cm) -- cycle;
			
			\draw[->,thick] (Center) -- ($(Center)!0.5cm!110:(B)$);
			
			\node[right] at ($(Center)!0.5cm!110:(B)$) {$\theta$};
			
		\end{tikzpicture}
		& &
		
		\begin{tikzpicture}[scale=0.63]
			
			\coordinate (O) at (0,0);

			\foreach \angle in {-10, 40, 90, 130, 220, 270} {
				\draw (O) -- (\angle:2.5);
			}
			\fill[red!5] (O) -- (-10:2.5) arc (-10:40:2.5) -- cycle;
			
			\node at (20:2) {$\calF_{y_1}$}; 
			\node at (65:2) {$\calF_{y_2}$};
			\node at (110:2) {$\calF_{y_3}$};
			\node at (180:2) {$\calF_{y_4}$};
			\node at (245:2) {$\calF_{y_5}$};
			\node at (-50:2) {$\calF_{y_6}$};
			
			\draw[->,thick] (O) -- (3:1.5);
			\node[right] at (3:1.5) {$\theta$};
			
		\end{tikzpicture} \\
		Normal cone to $y_1$ && Normal fan of $\conv(\calY)$
	\end{tabular}
	\caption{Normal cone to $\bar y$ and normal fan of $\calC$. Oracle $\hat y$ is piecewise constant on each cone of the normal fan.}
	\label{fig:piecewise_constant}
\end{figure}
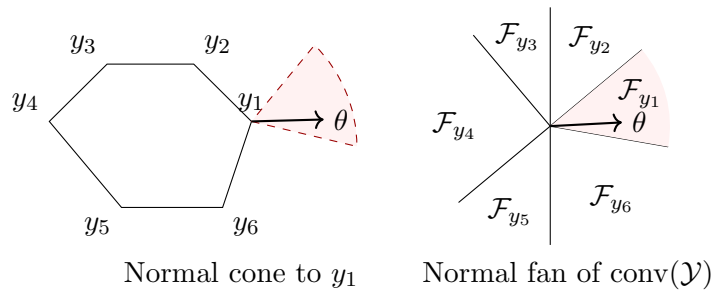
\begin{equation}\label{eq:Fenchel_Young_loss}
    \begin{aligned}
        \fyl_\Omega(\theta,\bar y) &= \max_{y \in \calC} \Big(\theta^\top y - \Omega(y)\Big) - \Big(\theta^\top \bar y - \Omega(\bar y) \Big)   \\
        &= \Omega^*(\theta) + \Omega(\bar y) - \theta^\top \bar y,
    \end{aligned}
\end{equation}
where $\Omega^*(\theta) = \max_{y \in \calC} \Big(\theta^\top y - \Omega(y)\Big)$ is the Fenchel conjugate of $\Omega$.
We note that the optimum in~\eqref{eq:regularized_combinatorial_optimization_oracle_polytope} is obtained in $\nabla \Omega^*(\theta)$, and that regularizing~\eqref{eq:combinatorial_optimization_oracle_polytope} amounts to replacing $\hat y$ by $\nabla \Omega^*$.

The loss $\fyl_\Omega$ has all the desirable properties of a loss in \gls{acr:ml}.
It is smooth, convex, and differentiable in $\theta$ with gradient:
\begin{equation}\label{eq:Fenchel_Young_loss_gradient}
    \nabla_\theta \fyl_\Omega(\theta,\bar y) = \nabla \Omega^*(\theta) - \bar y.
\end{equation}

This loss actually nicely captures the geometry induced by~\eqref{eq:regularized_combinatorial_optimization_oracle_polytope}.
Given a smooth and strictly convex $\Omega$, the \emph{Bregman divergence} associated to $\Omega$ is defined as:
\begin{equation}\label{eq:Bregman_divergence}
        \calD_\Omega(\bar y|y) = \Omega(\bar y) - \big(\Omega(y) + \nabla \Omega(y)^\top (\bar y - y)\big).
\end{equation}
It is the difference at $\bar y$ between the value of $\Omega$ and the value of the tangent of $\Omega$ at $y$, and thus measures the difference between $\bar y$ and $y$. It is a generalization of the squared Euclidean distance, which is obtained when $\Omega(y) = \frac{1}{2}\|y\|^2$.
Under the additional assumption that $\Omega$ is Legendre-type, the gradient of $\Omega$ induces a bijection from $\calC$ to $\bbR^d$, whose inverse is $\nabla \Omega^*$. If in addition $\calC$ is plain dimensional, \citet{blondelLearningFenchelYoungLosses2020} have shown that, given $y$ and $\bar y$ in $\calC$, their duals $\theta = \nabla \Omega^*(y)$ and $\bar \theta = \nabla \Omega^*(\bar y)$, we have:
\begin{equation}\label{eq:Bregman_divergence_Fenchel_Young_loss}
    \fyl_\Omega(\theta,\bar y) = \calD_\Omega(\bar y|y) = \calD_{\Omega^*}(\theta | \bar \theta).
\end{equation}
In other words, the \emph{\gls{acr:fy}-loss} is a primal-dual Bregman divergence associated to the linear oracle and:
$$ \fyl_\Omega\big(\varphi_w(x),\bar y\big) = D_{\Omega}\big(\bar y\big|\overbrace{\nabla\Omega^*\circ\varphi_w(x)}^{\substack{
    \text{Regularized} \\
    \text{neural network} \\
    \text{prediction}
}}\big). $$
It remains to define regularization functions $\Omega$ that lead to good prediction performance and tractable gradients. One natural choice is $\Omega(y) = \frac{1}{2}\|y\|^2$. The regularized prediction~\eqref{eq:regularized_combinatorial_optimization_oracle_polytope} then amounts to minimizing the squared Euclidean distance between the prediction and the target solution. 
It is a convex quadratic optimization problem that is tractable for some polytope~$\calC$. However, on some~$\calC$, only the non-regularized linear optimization oracle $\hat y$ is available. To overcome this obstacle, we introduce an $\Omega$ that requires only calls to~$\hat y$ in the following.

\paragraph{Entropic regularization.}
The negentropy is perhaps the regularization that is the most frequently used in the \gls{acr:ml} literature. 
It leads to a natural interpretation of the output as a probability distribution over $\calY$, this distribution belonging to the exponential family~\citep{wainwright2008graphical}. 
Minimizing the \gls{acr:fy}-loss amounts to compute the maximum likelihood estimator for this distribution, and we obtain the Kullback-Leibler divergence as Bregman divergence.
The exact computation of $\nabla \Omega^*(\theta)$ can however be challenging, unless $\max_{y\in\calC}\theta^\top y$ can be computed by dynamic programming~\citep{mensch2018differentiable}.
Stochastic gradients can however be easily obtained by Markov chain Monte Carlo as soon as a local search is available for $\max_{y\in\calC}\theta^\top y$ \citep{vivier2025learning}.

\paragraph{Perturbation}

Let us define the perturbed linear prediction: 
\begin{equation}\label{eq:perturbed_linear_prediction}
    F(\theta) = \bbE\Big[\argmax_{y \in \calC} (\theta + \varepsilon Z)^\top y\Big],
\end{equation}
where $\varepsilon > 0$ holds and $Z$ is a standard Gaussian vector on $\bbR^d$.
\citet{berthetLearningDifferentiablePerturbed2020} suggest using the Fenchel dual of $F$ as regularization $\Omega$.
They show that $F$ is smooth and strictly convex, and that $\Omega$ is Legendre-type, hence $F$ is its own bidual and $\Omega^* = F$. Danskin's theorem then gives:
$$\nabla \Omega^*(\theta) = \nabla F(\theta) = \bbE\Big[\argmax_{y \in \calY} (\theta + Z)^\top y\Big]. $$
Unbiased estimates of $\nabla \Omega^*$ and hence of $\nabla_\theta\fyl_\Omega(\theta,\bar y)$ can be obtained by sampling $Z$ and solving the resulting inner maximization problem using $\hat y$.
This gives a convenient way to solve the learning problem~\eqref{eq:supervised_learning_primal_dual_loss} via stochastic gradient descent using only calls to $\hat y$ to compute the stochastic gradients.

\subsection{Empirical Cost Minimization}
\label{sec:risk_minimization}
\Gls{acr:ecm} plays a central role in \gls{acr:coaml}, yet designing efficient algorithms for this setting is highly challenging. 
The objective is piecewise constant, which complicates gradient-based optimization, and existing approaches come with various limitations. 
For instance, black-box global optimization solvers can demonstrate that cost minimization yields excellent policies $\pi_w$ \citep{parmentierLearningStructuredApproximations2021}, but they are not compatible with deep learning and remain restricted to \glspl{acr:glm} of moderate dimension. 
Structured Hinge losses \citep{elmachtoubSmartPredictThen2021} provide a way to integrate cost minimization into deep architectures, but lead to tractability issues when the cost function $\fh$ is non-linear in $y$. 
Designing scalable and deep learning compatible cost minimization algorithms is therefore an important open problem for \gls{acr:coaml}. 
It is an active and rapidly growing research field, where new techniques are emerging to address the inherent difficulties. 

In the following, we review two approaches that constitute some of the first systematic attempts in this direction. 
The first is an alternating minimization method in the \gls{acr:cso} setting, which decomposes the learning problem into tractable substeps. 
The second is \gls{acr:srl}, which integrates \gls{acr:co}-layers into actor-critic architectures and stabilizes training through \gls{acr:fy} losses. 
Both approaches illustrate current directions to make cost minimization scalable and compatible with deep learning, while highlighting open challenges that remain central to \gls{acr:coaml}.

\paragraph{Altenating minimization algorithm in the \gls{acr:cso} setting}

A recent line of research by \citet{bouvier2025primal} proposes a generic \emph{alternating minimization algorithm} for combinatorial \gls{acr:cso}. This setting considers a decision-maker facing uncertainty $\bfxi \in \Xi$, while having access to a correlated context $x \in \mathcal{X}$. Given $x$, the policy $\pi$ selects a feasible decision $y \in \mathcal{Y}(x)$. The objective is to learn a conditional distribution $\pi(y|x)$ within a hypothesis class $\mathcal{H}$ that minimizes the expected cost
\begin{equation}
    \min_{\pi \in \mathcal{H}} R(\pi) 
    \quad \text{where} \quad 
    R(\pi) = \mathbb{E}_{(\bfx,\bfxi),\, \bfy \sim \pi(\cdot|x)}[c(\bfx,\bfy,\bfxi)].
\end{equation}

Since the joint distribution of $(x,\bfxi)$ is unknown, the learner only observes a dataset $\{(x_i,\xi_i)\}_{i=1}^N$. Policies are parameterized through a statistical model $\varphi_w$ that predicts a score vector $\theta = \varphi_w(x)$, which in turn parametrizes a \gls{acr:co}-layer returning a distribution over $\mathcal{Y}(x)$. Following the \gls{acr:fy} framework \citep{blondelLearningFenchelYoungLosses2020}, one defines for each context a regularized prediction layer:
\begin{equation}
    \pi_w(y|x) = p_{\Omega_{\mathcal{Y}(x)}}\!\left(y \,\middle|\, \varphi_w(x)\right) = \argmax_{q \in \Delta^{\calY(x)}} \big(\varphi_w(x)\big)^\top Y q -\Omega_{\calY(x)}(q) ,
\end{equation}
where $\Omega_{\calY(x)}$ is a convex regularization on the distribution simplex $\Delta^{\calY(x)}$ on $\calY(x)$, and $Y = (y)_{y \in \calY}$ is the wide matrix whose columns are the $y$ in $\calY$. Classical negentropy yields a softmax-type distribution, while the sparse perturbation regularizer of \citet{berthet2020learning} provides more general stochastic smoothing.
 
The \gls{acr:ecm} problem
\[
\min_{w} R_N(\pi_w) 
= \frac{1}{N}\sum_{i=1}^N \mathbb{E}_{\bfy \sim \pi_w(\cdot|x_i)}[c(x_i,\bfy,\xi_i)],
\]
is highly non-convex and suffers from high-variance gradients. To address this, \cite{bouvier2025primal} introduce a \emph{surrogate objective} coupling per-sample solutions with the model parameters:
\begin{equation}
    S(w,q_\otimes) = \frac{1}{N}\sum_{i=1}^N \Big( \bbE_{\bfy \sim q_i}\big[c(x_i,\bfy,\xi_i)\big] + \kappa L_{\Omega_\calY(x_i)}(Y_i^\top\varphi_w(x_i),q_i)\Big),
\end{equation}
where $q_\otimes=(q_1,\ldots,q_N)$ is a set of candidate distributions $q_i$ over $\Delta^{\calY(x_i)}$, $Y_i$ is the wide matrix whose columns are the $y$ for $y \in \calY(x_i)$, and $L_{\Omega_{\calY(x)}}$ is the \gls{acr:fy}-loss. The constant $\kappa>0$ controls the balance between data fidelity and regularization.
 
The core idea is to minimize $S$ by alternating between two tractable subproblems:
\begin{align}
    q_\otimes^{(t+1)} &\in \arg\min_{q_\otimes} S(w^{(t)},q_\otimes), 
    \quad \text{(decomposition step)} \label{eq:decompositionStep}\\
    w^{(t+1)} &\in \arg\min_{w} S(w,q_\otimes^{(t+1)}), 
    \quad \text{(coordination step)}.
\end{align}

The decomposition step separates across training samples and reduces to solving deterministic single-scenario problems, while the coordination steps amounts to \gls{acr:sl} using a \gls{acr:fy}-loss.
However, as such, they look intractable as both are on the space of distributions in $\Delta^{\calY(x)}$.

The key insight is that, using the appropriate regularization $\Omega_{\Delta^{\calY(x)}}$ on $\Delta^{\calY(x)}$, all the computations can be done on $\calC(x)$.
As in \gls{acr:sl}, the two practical regularizations are the entropic regularization and the regularization by perturbation.
When using the latter, we obtain an iteration of the form:
\begin{align}
    \mu_i^{(t+1)} &= \mathbb{E}_{\bfZ \sim \calN(0,\mathrm{Id})}\!\left[ \argmin_{y \in \calY(x_i)}\big(c(x_i,y,\xi_i\big) - \kappa \big(\varphi_{w^{(t)}}(x_i)+\bfZ)^\top y\right], \\
    w^{(t+1)} &= \arg\min_{w} \frac{1}{N}\sum_{i=1}^N L_{\Omega_{\calC(x_i)}}\!\big(\varphi_w(x_i),\mu_i^{(t+1)}\big).
\end{align}
 Obtaining a Monte Carlo estimate of the decomposition step only requires to solve a deterministic single scenario problem of the form:
\begin{equation}
    \label{eq:singleScenarioOptim}
    \min_{y \in \mathcal{Y}(x_i)} \, c(x_i,y,\xi_i) + \langle \theta, y \rangle,
\end{equation}

which is assumed tractable given an oracle for the deterministic problem. 
The coordination step becomes a \gls{acr:sl} problem using a \gls{acr:fy}-loss with a regularization by perturbation on $\calC(x_i)$: this is the usual \gls{acr:sl} setting for \gls{acr:coaml} and can be efficiently solved using stochastic gradient descent.
In fact, this structure makes the method scalable: heavy combinatorial solves are done offline within the decomposition step, while the coordination step leverages standard stochastic gradient methods. Using negentropy as regularizer leads to an inference problem in an exponential family for the first step, and a maximum likelihood estimator in the second.

Conceptually, the decomposition step leverages existing efficient solvers for deterministic problems, while the coordination step leverages gradient-based updates typical for deep learning. 
This alternating structure allows the algorithm to exploit the best of both worlds: powerful \gls{acr:co}-oracles on the one hand, and scalable learning updates on the other. Moreover, this procedure inherits desirable properties: (i) it is compatible with deep learning frameworks through automatic differentiation; (ii) it is generic across combinatorial structures provided single-scenario problems are tractable; (iii) it admits theoretical convergence guarantees under mild assumptions, linking to mirror descent and Bregman geometry; and (iv) empirically, it matches the performance of computationally heavy fully-coordinated heuristics, e.g., Lagrangian relaxations, at a fraction of the computational cost. Notably, the method highlights how coordination across scenarios---achieved via alternating minimization---is essential to go beyond purely anticipative or uncoordinated imitation strategies.

\paragraph{Structured Reinforcement Learning (\gls{acr:srl})}
\gls{acr:rl} has shown impressive success in robotic control and gaming, but its application to large-scale industrial problems is hindered by the combinatorial structure of action spaces. Standard actor-critic methods such as proximal policy optimization or soft-actor-critic struggle in these settings, as they need to represent distributions over exponentially many feasible actions, and cannot easily exploit combinatorial structure. To overcome this challenge, \citet{hoppeStructuredReinforcementLearning2025} propose \emph{\gls{acr:srl}}, a new actor-critic framework that integrates \gls{acr:co} layers into the actor policy and employs \gls{acr:fy}-losses to enable stable end-to-end training.
 
\gls{acr:srl} uses the \gls{acr:coaml} architectures introduced in Section~\ref{sec:multistageStoOpt} for the actor. To maintain consistency with the rest of this paper, we denote the state by $x$ and the action by $y$, deviating from the standard RL notation $s$ and $a$.
A statistical model $\varphi_w$ encodes the state $x$ into a score vector $\theta = \varphi_w(x)$, which parametrizes a combinatorial optimizer $\hat y$. The \gls{acr:co}-layer then computes the action
\begin{equation}
    y = \hat y(\theta,x) \in \argmax_{\tilde y \in \calY(x)} \langle \theta, \tilde y \rangle,
\end{equation}
ensuring that every selected action $y$ is feasible with respect to the problem-specific constraints defining~$\calY(x)$. The resulting actor policy is $\pi_w(\cdot|x) = \delta_{\hat y(\varphi_w(x),x)}$, i.e., a Dirac measure on the action returned by the \gls{acr:co}-layer. This structure guarantees feasibility by design and scales to very large action spaces.

Like the other methodologies, \gls{acr:srl} must deal with the \gls{acr:coaml} challenges. While inference is straightforward, learning the actor is highly non-trivial: the \gls{acr:co}-layer is piecewise constant with respect to $\theta$, producing zero gradients almost everywhere. 

The \gls{acr:srl} builds upon the \gls{acr:sl} and \gls{acr:ecm} methodolgies to resolve these issues.
From \gls{acr:sl}, it borrows entropic regularization and the regularization by perturbation to smooth the actor, and the FY-loss to train it.
From \gls{acr:ecm}, it takes the alternating minimization idea.
However, this lets open two challenges. The exploration, as we cannot follow the expert as in the imitation learning approach to multistage problems.
This is done through standard rollout.
The second is that the decomposition step~\eqref{eq:decompositionStep} is doubly intractable: First because we do not have access to a model and to full information on $\bfxi$ in a \gls{acr:rl} setting, and therefore for not have an oracle for~\eqref{eq:singleScenarioOptim}. 
This is resolved in two ways: we use a critic to turn the multistage problem~\eqref{eq:singleScenarioOptim} into a single stage problem, and instead of solving it on the full $\calY(x)$, we just sample a few decisions from the actor and evaluate them thanks to the environment and the critic.

\begin{figure}[!hb]
	\centering
	\includegraphics[width=\textwidth]{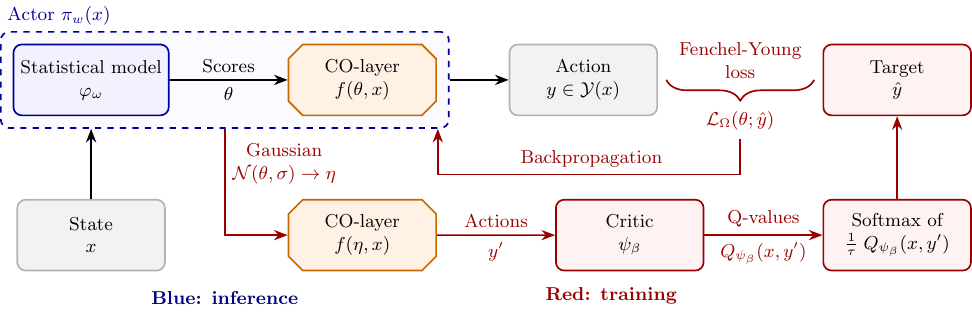}
	\caption{Overview of the components of \gls{acr:srl}.}
	\label{fig:srl_algorithm}
\end{figure}

Figure~\ref{fig:srl_algorithm} visualizes the respective training pipeline.
At each update step, the actor first computes a score vector $\theta = \varphi_w(x)$ from the current state.
The algorithm samples Gaussian perturbations $\eta \sim \mathcal{N}(\theta,\sigma_b)$ and passes each perturbed score vector through the \gls{acr:co}-layer to produce a feasible candidate action $y'=\hat y(\eta,x)$.
The critic $Q_\psi(x,y')$ evaluates the quality of these candidates, and we apply a softmax aggregation with temperature parameter $\tau$ to assign weights to them.
This aggregation yields a convex combination $\bar y$ that represents a smooth target action lying in the convex hull of feasible actions.
Finally, we update the actor parameters by minimizing a \gls{acr:fy}-loss $L_\Omega(\theta;\bar y)$, which provides stable gradients, while updating the critic in parallel using standard temporal-difference learning.
In this way, \gls{acr:srl} sidesteps the non-differentiability of the \gls{acr:co}-layer and instead propagates informative gradients through smooth surrogate targets constructed from critic evaluations.

The main loop shown in Algorithm~\ref{alg:srl} proceeds as follows.
The actor encodes the state $x$ into scores $\theta=\varphi_w(x)$, which it perturbs to enhance exploration. The combinatorial oracle utilizes these perturbed scores to generate actions $y=\hat y(\eta,x)$.
We store the resulting transitions $(x,y,r,x')$ in a replay buffer to allow off-policy learning. During the actor update, the critic $Q_\psi$ evaluates candidate actions resulting from multiple perturbations around $\theta$, and a softmax aggregation of these evaluations yields a convex target action $\bar y$.
We then update the actor by minimizing a \gls{acr:fy}-loss, which provides smooth gradients even though the \gls{acr:co}-layer itself is non-differentiable.
The critic update follows standard temporal-difference learning by regressing the Q-function toward a target that includes rewards and bootstrapped value estimates. In this way, \gls{acr:srl} combines feasibility-preserving action selection using an oracle with stable actor learning through \gls{acr:fy}-surrogates, while keeping the overall training loop structurally similar to standard actor-critic methods.

Geometrically, \gls{acr:srl}’s actor step can be viewed as a projection onto the moment polytope $\calC(x)=\mathrm{conv}(\calY(x))$: the critic induces a preference over extreme points (feasible actions), the softmax forms a point $\bar y \in \calC(x)$, and the \gls{acr:fy}-update adjusts $\theta$ to increase the support of high-value regions of $\calC(x)$ without crossing non-differentiable cone boundaries. This yields low-variance and decision-aware gradients while preserving feasibility throughout training. 
The overall result
\begin{algorithm}[!ht]
	\small
	\caption{Structured Reinforcement Learning}\label{alg:srl}
	\begin{algorithmic}[1]
		\State Initialize actor $\varphi_w$, critic $Q_\psi$, and target critic $\bar Q_\psi$
		\For{each episode}
		\State Collect transitions $(x,y,r,x')$ utilizing $\eta \sim \mathcal{N}(\varphi_w(x),\sigma_f)$ and $y=\hat y(\eta,x)$
		\State Store transitions in replay buffer $\mathcal{D}$
		\For{each sampled batch $(x,y,r,x') \sim \mathcal{D}$}
		\State \textbf{Actor update:} 
		\begin{enumerate}[leftmargin=2.2cm]
			\item Sample $m$ perturbations $\eta$ around $\theta=\varphi_w(x)$
			\item Compute candidate actions $y'=\hat y(\eta,x)$ and Q-values $Q_\psi(x,y')$
			\item Compute target action $\bar y$ from softmax over Q-values
			\item Update $w$ by minimizing \gls{acr:fy}-loss $L_\Omega(\theta;\bar y)$
		\end{enumerate}
		\State \textbf{Critic update:} Minimize TD-error $(Q_\psi(x,y)-[r+\gamma \bar Q_\psi(x',y')])^2$
		\EndFor
		\EndFor
	\end{algorithmic}
\end{algorithm}
 is a stable learning loop that couples powerful combinatorial inference through $\hat y$ with smooth optimization through \gls{acr:fy}-losses, enabling actor-critic training at combinatorial scale.

In most single-stage industrial problems, generating a good solution is more difficult than evaluating a given solution if the cost function is known. Often, obtaining optimal solutions for such problems is computationally infeasible. \gls{acr:srl} utilizes this asymmetry by repeatedly proposing candidate solutions and evaluating them to estimate an approximate target action. In single-stage settings, \gls{acr:srl} thus never requires solving the hard optimization problem, but only solving the simpler optimization in the oracle and evaluating the generated solutions. In multi-stage settings, this translates to using a critic to evaluate candidate actions generated by the combinatorial actor. In these settings, it is not necessary to estimate the anticipative offline problem.

\gls{acr:srl} combines the feasibility guarantees of \gls{acr:co}-layers with the scalability of neural function approximation. It requires no expert demonstrations (unlike structured \gls{acr:il}), and outperforms unstructured \gls{acr:rl} by large margins in environments with combinatorial action spaces. Its benefits include: (i) feasibility of actions by construction; (ii) reduced gradient variance through \gls{acr:fy}-updates; (iii) improved exploration via perturbation-based sampling; and (iv) strong empirical performance across static and dynamic benchmarks. The main trade-off is computational: the actor update involves multiple calls to the \gls{acr:co}-layer per step, making training slower than standard actor-critic learning or structured \gls{acr:il}. Nevertheless, \gls{acr:srl} provides a general paradigm for \gls{acr:rl} in \glspl{acr:mdp} with combinatorial action spaces, unifying structured prediction and actor-critic training.

\section{From Theory to Practice: A Guide to Applying \gls{acr:coaml}}
With the core components of \gls{acr:coaml} established, we now turn to their practical implementation.
While specific requirements naturally vary by application, effective deployment relies on a principled approach to two fundamental design phases: architecture selection and learning strategy.
Architectural choices are primarily dictated by the problem's underlying geometry and the available data modalities.
The learning strategy, in contrast, is governed by the interplay of three factors: the nature of the uncertainty, the temporal structure of the decision problem (static vs.\ dynamic), and the computational scale.
Table~\ref{tab:methodology_comparison} summarizes the intrinsic trade-offs between \gls{acr:sl}, \gls{acr:ecm}, and \gls{acr:srl}.
However, the feasible set of algorithms is fundamentally constrained by the uncertainty model—specifically, whether it is explicit or implicit.
The temporal dimension further refines this choice: dynamic settings inherit the principles of static counterparts but impose the additional burdens of state exploration and long-term credit assignment, requiring pipelines that tightly integrate predictive modeling with sequential optimization.
Finally, computational scale acts as a practical filter, often ruling out approaches that require excessive oracle calls.
In the following, we organize our guidelines along these axes, beginning with a taxonomy of uncertainty models before detailing specific learning strategies for static and dynamic environments.
\begin{table}[!ht]
	\centering
	\footnotesize
	\caption{Comparative analysis of the three primary methodological paradigms in \gls{acr:coaml}.}
	\label{tab:methodology_comparison}
	\begin{tabular}{p{2.5cm} p{3.2cm} p{3.2cm} p{4.5cm}}
		\toprule
		\textbf{Paradigm} & \textbf{Required Supervision} & \textbf{Gradient Source} & \textbf{Key Trade-off} \\
		\midrule
		\textbf{Empirical cost min. (\gls{acr:ecm})} & 
		Historical Contexts and Noises $(x_i, \xi_i)$ \newline Cost function $c$ & 
		Chain rule via implicit differentiation or surrogates & 
		\textbf{Pros:} Unbiased w.r.t.\ decision cost, perf.~and gen.~ guarantees. \newline 
		\textbf{Cons:} Best alg.~require linear obj.~or single scen.~oracle. \\
		\addlinespace[0.5em]
		\textbf{Imitation learning (\gls{acr:il})} & 
		Contexts $x_i$ \newline Expert Decisions $y_i^*$ & 
		Distance to expert target (e.g., Cross-Entropy, \gls{acr:fy}) & 
		\textbf{Pros:} Stable, fast training. \newline 
		\textbf{Cons:} Bounded by expert quality. \\
		\addlinespace[0.5em]
		\textbf{Reinforcement learning (\gls{acr:rl})} & 
		Interaction with environment \newline Scalar Reward $r$ & 
		Perturbed critics or surrogate targets (FY) & 
		\textbf{Pros:} Works with implicit dynamics. \newline 
		\textbf{Cons:} High sample complexity. \\
		\bottomrule
	\end{tabular}
\end{table}

Before diving into specific strategies, we offer a fundamental piece of practical advice: always begin by training a simple \gls{acr:glm} scoring model using the \gls{acr:fy} loss.
This recommendation stands even if it necessitates imitating a suboptimal synthetic expert or intentionally overfitting a single data point.
The rationale is grounded in convexity: unlike deep neural networks, this specific formulation yields a convex optimization problem where convergence is theoretically guaranteed.
This property transforms the initial training run into a reliable diagnostic tool.
If the model fails to learn, the failure can be attributed directly to bugs in the \gls{acr:co}-layer implementation rather than the instability of the training dynamics.
This step thus serves to validate the overall approach before introducing the complexities of deep architectures.
The guidelines that follow assume this foundational check has been successfully established.

\subsection{Architecture Design Guidelines}
The design of the \gls{acr:co}-layer serves as the cornerstone of the entire architecture.
As is common in \gls{acr:or}, this choice is heavily influenced by the scalability of available combinatorial oracles for the specific domain, often requiring domain knowledge and craftsmanship to select the appropriate elements that need to be modeled or approximated.
Crucially, this choice constrains the statistical encoder $\varphi_w$: the network must produce a parameter vector $\theta$, e.g., edge weights or node utilities, that matches the input interface of the chosen solver.
Once the \gls{acr:co}-layer is fixed and validated using the linear baseline recommended above, one should select a richer architecture that aligns with the structural dependencies of the data.
Table~\ref{tab:architectures_summary} summarizes the main architectural families; we elaborate on the rationale for each below.

\paragraph{Independent Scoring (GLM, MLP).}
The simplest approach models the parameters $\theta$ as independent functions of the context features.
For instance, in a vehicle routing problem, the cost of an edge might be predicted solely based on the distance and traffic attributes of that specific edge, ignoring the state of neighboring arcs.
\glspl{acr:glm} or simple \gls{acr:mlp} are the standard choices here.
While they ignore the complex interdependencies between decision variables, they offer significant advantages in terms of scalability and interpretability.
Furthermore, as noted earlier, \gls{acr:glm} encoders combined with \gls{acr:fy} losses yield convex optimization problems, making them the ideal starting point for any implementation.

\paragraph{Graph Neural Networks (GNNs).}

Many combinatorial problems, such as routing, flow, or transportation, are naturally defined on graph structures.
In these settings, independent scoring is often insufficient because the "value" of a node or edge depends heavily on its topological neighbors.
Graph Neural Networks explicitly encode this relational structure, passing messages between nodes to compute context-aware embeddings.
This allows the encoder to capture local dependencies—such as congestion propagating through a network—before the solver is even invoked.
While computationally more demanding than independent models, GNNs are typically required to achieve state-of-the-art performance on network-structured tasks.
\begin{table}[!ht]
	\centering
	\caption{Summary of statistical encoder $\varphi_w$ architectures for \gls{acr:coaml}.}
	\label{tab:architectures_summary}
	\small
	\begin{tabular}{p{3cm}p{4cm}p{3.5cm}p{3.5cm}}
		\toprule
		\textbf{Architecture} & \textbf{Typical Applications} & \textbf{Pros} & \textbf{Cons} \\
		\midrule
		\textbf{Independent Scoring} \newline (GLM, MLP) & Learning heuristics for hard static CO (e.g., SVSP, SMSP). & Convex \gls{acr:sl} for GLM, scalable, simple to implement, interpretable. & Ignores structural dependencies between decision variables. \\
		\addlinespace[0.5em]
		\textbf{Graph Neural Networks} \newline (GNN) & Network-structured problems (e.g., VRP, flows, fixed-charge transportation). & Captures relational structure and local dependencies naturally. & Higher computational cost; can be harder to train than independent models. \\
		\addlinespace[0.5em]
		\textbf{Autoregressive / Sequential} \newline (RNN, Transformer) & Dynamic and multi-stage settings (e.g., routing w/ time windows, packing). & Explicitly models sequential decision-making and history. & Inference can be slow; training often requires specific techniques (e.g., teacher forcing). \\
		\addlinespace[0.5em]
		\textbf{Domain-Informed} \newline (CNN, PINN-like) & Contextual Stochastic Optimization (e.g., inputs are images, sensor data). & Leverages prior physical/spatial knowledge; high sample efficiency. & Highly problem-specific; requires custom design. \\
		\bottomrule
	\end{tabular}
\end{table}

\paragraph{Autoregressive and Sequential Models.}
For dynamic settings or multi-stage problems, the temporal dimension becomes critical.
Here, the decision at time $t$ depends not just on the current context, but on the history of past states and actions.
Architectures such as \glspl{acr:rnn} or Transformers are designed to encode these sequential dependencies into a hidden state representation.
These models are particularly relevant for applications like dynamic bin packing or online routing, where the "context" is an evolving stream of requests rather than a static snapshot.

\paragraph{Domain-Informed Architectures.}
Finally, in Contextual Stochastic Optimization, the input $x$ often consists of raw, unstructured data such as satellite imagery for routing or sensor readings for scheduling.
In these cases, the architecture must bridge the gap between perception and reasoning.
\glspl{acr:cnn} are standard for spatial inputs, while \glspl{acr:pinn} can be used to respect underlying physical laws, e.g., in energy grid management.
These architectures leverage strong inductive biases to extract meaningful cost parameters $\theta$ from high-dimensional inputs with high sample efficiency.


\subsection{Learning Guidelines: Taxonomy of Uncertainty}

We structure the selection of learning algorithms around the nature of the uncertainty, which serves as the primary determinant for the appropriate methodology.
Building on the formalization in Section~\ref{sec:formalizationApplication}, we frame this discussion within the \gls{acr:cso} model, as it provides a unified abstraction for the various static settings.
Within this framework, we distinguish three canonical uncertainty regimes, classified by the observability and explicitness of the random parameters.

\paragraph{Explicit and fully observable uncertainty.}
    Here, the random variable $\bfxi$ has a known representation and realizations can be observed directly. 
    The optimization problem is then $\min_{y \in \mathcal{Y}(x)} \mathbb{E}[c(x,y,\bfxi)]$, with $c$ being tractable once $\bfxi$ is revealed. 
    In this case, purely model-free approaches are possible: one can optimize directly against realizations of $\bfxi$, which corresponds to fully decision-focused~\gls{acr:sl}.
\paragraph{Explicit but partially observable uncertainty.}
    The random variable $\bfxi$ has a known representation, but only indirect or noisy signals are observed, e.g., downstream costs or partial state transitions. 
    This requires an estimate-then-optimize pipeline, where statistical learning reconstructs $\bfxi$ or its impact before optimization. 
    This setting is common in practice, notably in pricing applications, but under-explored from a methodological point of view.
\paragraph{Implicit uncertainty.}
    In some problems, the uncertainty distribution is not available in closed form. 
    Instead, one only has access to a simulator or sampling oracle, i.e., $\mathbb{E}[c(x,y,\bfxi)]$ or $\mathbb{P}[s_{t+1}|s_t,y_t]$ cannot be written explicitly. 
    Even if representer theorems suggest the existence of a latent $\bfxi$, it cannot be integrated into a tractable optimization model. 
    This setting connects \gls{acr:coaml} to \gls{acr:rl}, since optimization must proceed through repeated interaction with the oracle.

\noindent Table~\ref{tab:learning_guidelines} summarizes the recommended learning approaches across these different regimes.

\paragraph{Supervised learning.}
Consider first the scenario where \emph{we have a training set $(x_i,\bar y_i)_i$ derived from an optimal or high-quality expert policy}, as illustrated in Example~\ref{ex:paths_in_images}.
In this context, the primary role of \gls{acr:coaml} is to distill this policy, typically because the expert is computationally too expensive for real-time application.
Here, the most effective strategy is to directly imitate the expert using a \gls{acr:fy}-loss.

\paragraph{Explicit and fully observable uncertainty.}
In the second setting, the \emph{uncertainty $\bfxi$ is explicit and fully observable}, as seen in Example~\ref{ex:stochastic_vsp}.
The training set comprises pairs $(x_i,\xi_i)_i$ rather than pre-computed decisions.
This setting is particularly favorable when the corresponding deterministic optimization problem
\begin{equation}
	\label{eq:anticipativeCO}
	\min_{y \in \mathcal{Y}(x)} \, c(x,y,\xi),
\end{equation}
is tractable once the realization $\xi$ of $\bfxi$ is known.
In this setting, a robust baseline strategy is to construct a supervised dataset $(x_i,\bar y_i)_i$ by solving the anticipative problem
$
\bar y_i=\min_{y \in \mathcal{Y}(x_i)} \, c(x_i,y,\xi_i)
$.
However, if $\bfxi$ exhibits high conditional variance given $x$, this anticipative solution may diverge significantly from the true optimal policy.
In such cases, coordination is needed to bridge the gap between the anticipative and stochastic optima, making \gls{acr:ecm} a superior choice.
Nevertheless, starting with \gls{acr:sl} remains sound practice, as it provides a stable initialization for subsequent \gls{acr:ecm} fine-tuning.
Conversely, if the ex-post problem~\eqref{eq:anticipativeCO} is itself intractable, one must resort to \gls{acr:srl}.

\paragraph{Explicit but partially observable uncertainty.}
Frequently, full realizations of $\bfxi$ are unavailable; instead, we observe only partial feedback, such as the realized cost $c(x_i,y_i,\xi_i)$ of a specific decision.
A central challenge in this setting is the inability to evaluate the counterfactual cost $c(x_i,y_i',\xi_i)$ of an alternative decision $y_i'$, which precludes direct optimization.
Current literature has only partially addressed this issue, leaving significant room for methodological innovation in handling noisy or censored observations of $\bfxi$.
At present, the prevailing approach is the \gls{acr:pto} paradigm: one first reconstructs the distribution of $\bfxi$ or its impact on $c$ via statistical estimation, and then applies the methodology for fully observable settings.
While this reduces to the previous case if the estimation is exact, the optimal integration of statistical inference and decision-focused optimization under partial observability remains a key open research question.
\begin{table}[!hb]
	\centering
	\caption{Recommended learning paradigms based on problem setting and uncertainty type.}
	\label{tab:learning_guidelines}
	\small
	\begin{tabular}{p{4cm}p{5.5cm}p{5.5cm}}
		\toprule
		\textbf{Scenario} & \textbf{Static Setting} & \textbf{Dynamic Setting} \\
		\midrule
		\textbf{Learning from Expert} \newline (Demonstrations available, training set of $(x,\bar y)$) & \textbf{Supervised Learning} \newline Minimize \gls{acr:fy}-loss against expert decisions. & \textbf{Imitation Learning} \newline Supervised learning on expert trajectories, augmented with off-policy techniques (e.g., DAgger) to address distribution shift. \\
		\addlinespace[0.5em]
		\textbf{Explicit \& Fully Observable uncertainty} \newline (Training set of $(x, \xi)$) & \textbf{1. SL on Anticipative Solutions} \newline Solve $\min_y c(x,y,\xi)$ to get targets, then SL. \newline \textbf{2. Empirical Cost Min.} \newline Alternating minimization for better coordination. & \textbf{1. IL on A Posteriori Optima} \newline Compute optimal trajectory given full $\xi_{1:T}$, then imitate. \newline \textbf{2. Differentiable SP} \newline (Open research question). \\
		\addlinespace[0.5em]
		\textbf{Explicit \& Partially Observable} \newline (Noisy/Partial signals) & \textbf{Estimate-then-Optimize} \newline Reconstruct $\xi$ or cost, then apply fully observable methods. (Decision-aware integration is open). & \textbf{Estimate-then-Optimize} \newline Largely unexplored in \gls{acr:coaml}. \\
		\addlinespace[0.5em]
		\textbf{Implicit Uncertainty} \newline (Simulator/Oracle only) & \textbf{SRL-inspired methods} \newline Use perturbation/gradient estimation techniques from \gls{acr:srl}. & \textbf{Structured Reinforcement Learning} \newline \gls{acr:srl} (Actor-Critic with \gls{acr:co}-layer and \gls{acr:fy}-updates). \\
		\bottomrule
	\end{tabular}
\end{table}

\paragraph{Implicit uncertainty.}
Finally, some problems fall into the category of \emph{implicit uncertainty}, where no explicit representation of $\bfxi$ is available. 
Instead, the learner only has access to a simulator or oracle for costs or transitions. 
Formally, one can query an oracle to query samples that allow to approximate $\mathbb{E}[c(x,y,\bfxi)|x,y]$
as well as state transition probabilities $\mathbb{P}[s_{t+1}|s_t,u_t]$, but these cannot be integrated into a tractable optimization model. 
This setting includes deterministic black-box optimization problems, where the cost function is itself an oracle with no closed form. 
In such cases, classical decision-focused learning is not directly applicable. 
Instead, methods inspired by \gls{acr:srl} are promising: even though the setting is static, one can treat the uncertainty as part of the environment and rely on perturbation-based updates or surrogate losses to propagate gradients through the \gls{acr:co}-layer. 
This connects static implicit problems to the \gls{acr:rl} techniques discussed in Section~3.3.


\subsection{Learning Guidelines: Dynamic Settings}\label{subsec:dynamicSetting}
Dynamic problems extend the static setting by introducing temporal structure: a sequence of states $(x_t)_{t=0}^T$ evolves according to state transitions $x_{t+1} = \calT(x_t,y_t,\xi_t)$ that depend on decisions $y_t$, random noise realizations $\xi_t$, and possibly contextual information $x_t$.
A policy must therefore map states and contexts to feasible actions, with the objective of minimizing expected cumulative cost.

\paragraph{Imitation learning and state space exploration.}
In the context of multistage stochastic optimization problems~\eqref{eq:multistage_stochastic_optimization}, \gls{acr:sl} is effectively \emph{imitation learning}: the policy $\pi_w$ is trained to mimic a target policy $\pi^*$.
The fundamental approach involves minimizing a Fenchel--Young loss over a dataset generated by $\pi^*$.
The primary limitation of this method is the issue of distributional shift: it minimizes the discrepancy between $\pi_w$ and $\pi^*$ in expectation over the states visited by the expert $\pi^*$, whereas the ideal objective is to minimize this discrepancy over the states visited by the learner $\pi_w$.
Consequently, if $\pi_w$ cannot perfectly approximate $\pi^*$—due to architectural constraints or limited information—the resulting policy may underperform, as it makes accurate decisions on states it rarely visits while failing on the states it actually encounters.
To mitigate this, one must employ off-policy learning techniques~\citep{greif2024combinatorial,rautenstrauss2025optimization} that account for the induced state distribution.

\paragraph{Explicit uncertainty and multistage policies to imitate.}
When uncertainty is \emph{explicit and fully observable}, the training data consists of complete episodes $\xi_1,\ldots,\xi_T$.
As in the static case, the most direct strategy is to derive an expert policy for imitation.
Since the full uncertainty realization is known a posteriori, computing the optimal trajectory $x_1,y_1,\ldots,x_T,y_T$ for the episode reduces to a deterministic problem that can be solved offline.
Let $\pi^*$ denote the corresponding policy.
Although this \emph{anticipative} policy relies on future information $\xi_{t+1},\ldots,\xi_T$ and cannot be deployed online, it serves as an ideal oracle for generating a training set $x_1,y_1,\ldots,x_n,y_n$.
Solving the \gls{acr:sl} problem~\eqref{eq:supervised_learning_primal_dual_loss} then yields a non-anticipative, deployable policy $\pi_w$ that imitates $\pi^*$ \citep{batyCombinatorialOptimizationEnrichedMachine2024}.
This remains the predominant approach in the literature.

It is worth noting that the anticipative policy represents the simplest form of multistage stochastic optimization.
Where computationally tractable, one may instead imitate more sophisticated policies, such as two-stage approximations or scenario trees.
Furthermore, rather than merely imitating a multistage policy, a promising research direction is to adapt the solution algorithms of these policies directly into learning algorithms capable of optimizing on-the-fly.
This remains an open and fertile area of research.
Similarly, the challenge of partially observable uncertainty in dynamic settings has received limited attention to date.

\paragraph{Implicit models and structured \gls{acr:rl}.}
In many practical scenarios, system or environment dynamics are accessible only via simulation or direct environmental interaction.
With explicit transition functions being unavailable, the problem falls within the domain of \gls{acr:rl}.
While classical \gls{acr:rl} methods such as Q-learning or actor-critic algorithms can be appliied in this setting, they typically fail to exploit the combinatorial structure of the action space, such that scalability becomes a rapid bottleneck.
\gls{acr:coaml} architectures address this by embedding \gls{acr:co} oracles directly into the policy: the actor network produces scores, and a \gls{acr:co}-layer selects feasible actions.
This design ensures that exploration strictly respects problem-specific constraints and avoids infeasible actions—a critical requirement for industrial applications.
As of today, the only algorithm that can train such architecture is  \gls{acr:srl} \citep{hoppeStructuredReinforcementLearning2025}, which we introduced in Section~\ref{sec:risk_minimization}.

\section{State of the Art}\label{sec:stateoftheart}
The literature on \gls{acr:coaml} has grown rapidly in recent years, spanning a wide range of problem settings, algorithmic paradigms, and applications. 
While the previous sections introduced the technical foundations of \gls{acr:coaml} pipelines, we now provide a structured overview of the state of the art. 

We survey the literature from three perspectives.
Section~\ref{sec: state of art application} surveys the literature from an application perspective, supported by detailed summary tables. Section~\ref{sec: state of art methodology} complements this view by analyzing \gls{acr:coaml} contributions from a methodological perspective, comparing \gls{acr:ecm}, \gls{acr:il}, and \gls{acr:rl} approaches. Lastly, Section~\ref{subsec:guarantees} gives a brief outlook on statistical learning guarantees.

Throughout this section, we rely on tables to present the large body of work in a compact and accessible way. 
Our aim is not to provide an exhaustive list of papers, but to distill the most important ideas, illustrate representative contributions, and identify recurring patterns and open challenges.

\subsection{Application Perspective} \label{sec: state of art application}
\gls{acr:coaml} pipelines have been applied across a broad spectrum of problems, ranging from classical hard \gls{acr:co} to multi-stage decision-making under uncertainty. Tables~\ref{tab:applications_static} and~\ref{tab:applications_dynamic} provide an overview by summarizing representative works in detail. 
Table~\ref{tab:applications_static} covers static settings, including hard \gls{acr:co} problems, \gls{acr:cso}, and two-stage stochastic programs. 
Table~\ref{tab:applications_dynamic} covers dynamic settings, including multi-stage stochastic optimization problems and multi-stage problems with continuous or large action spaces. 
For each entry, we report the application, the learning method, the employed loss function, the statistical model, and the associated \gls{acr:co}-layer. 
This structured presentation reveals which methodological choices dominate in each application area.
\begin{table}[!hb]
	\footnotesize
	\centering
	\begin{threeparttable}
		\caption{Representative literature on \gls{acr:coaml} pipelines in static settings}
		\label{tab:applications_static}
		\setlength{\belowcaptionskip}{0.3cm}
		\setlength\tabcolsep{3pt}
		\begin{tabular}{p{2.2cm} p{2.4cm} p{1.4cm} p{3cm} p{2.2cm} p{3cm}}
			\toprule
			\textbf{Reference} & \textbf{Application} & \textbf{Method} & \textbf{Loss function} & \textbf{Stat. model} & \textbf{\gls{acr:co}-layer} \\
			\midrule
			\multicolumn{6}{c}{\textbf{Hard combinatorial optimization problems}} \\
			\hline
			\cite{parmentierLearningApproximateIndustrial2021} & Path problems, \gls{acr:svsp} & \gls{acr:sl} & Log-likelihood & \gls{acr:nn} & Relaxed \gls{acr:vsp} \\
			\cite{parmentierLearningStructuredApproximations2021} & \gls{acr:svsp}, \gls{acr:smsp} & \gls{acr:il}/\gls{acr:ecm} & \gls{acr:fy} / perturbed & \gls{acr:glm} & \gls{acr:vsp}, $1||\sum_j C_j$ \\
			\cite{dalle2022learning} & Warcraft, \gls{acr:svsp}, \gls{acr:smsp}, 2SMWST & \gls{acr:il}/\gls{acr:ecm} & \gls{acr:fy} / empirical regret & \gls{acr:nn}, \gls{acr:glm} & Shortest path, \gls{acr:vsp}, spanning forest \\
			\cite{parmentier2023structured} & \gls{acr:smsp} & \gls{acr:sl} & \gls{acr:fy} & Linear model & $1||\sum_j C_j$ \\
			\cite{aubin2024generalization} & \gls{acr:svsp} & \gls{acr:ecm} & Empirical regret & \gls{acr:glm} & \gls{acr:vsp} \\
			\cite{spieckermann2025reduce} & \gls{acr:fctp} & \gls{acr:il} & Cross-entropy & \gls{acr:gnn} & Reduced \gls{acr:fctp} \\
			\cite{wilder2019End} & Graph partition, node selection & \gls{acr:il} & Surrogate loss & Clustering layer & Partition relaxations \\
			\cite{berthet2020learning} & Shortest path & \gls{acr:il}/\gls{acr:ecm} & \gls{acr:fy} / empirical regret & ResNet & Shortest path \\
			\hline
			\multicolumn{6}{c}{\textbf{Contextual Stochastic Optimization problems}} \\
			\hline
			\cite{lodi2020learning} & Facility location & \gls{acr:il} & \gls{acr:mse} & Tree, \gls{acr:nn}, Naive Bayes, LogReg & Facility location \\
			\cite{greif2024combinatorial} & \gls{acr:dirp} & \gls{acr:sl} & \gls{acr:fy} & \gls{acr:pinn} & Prize-collecting \gls{acr:tsp} \\
			\cite{jungel2024wardropnet} & Traffic equilibria & \gls{acr:il} & \gls{acr:fy} & \gls{acr:nn}, \gls{acr:gnn} & Wardrop equilibrium \\
			\cite{bouvier2025primal} & Contextual 2SMWST & \gls{acr:ecm} & \gls{acr:fy} & \gls{acr:nn} & Spanning forest \\
			\cite{kong2022end} & Power scheduling, security games & \gls{acr:il} & Log-likelihood & \gls{acr:nn}, GRU & Surrogate problems \\
			\hline
			\multicolumn{6}{c}{\textbf{Two-stage stochastic optimization problems}} \\
			\hline
			\cite{bengio2020learning} & 2SIP & \gls{acr:il} & \gls{acr:mse} & Linear regression, \gls{acr:nn} & Deterministic MIP \\
			\cite{dalle2022learning} & 2SMWST & \gls{acr:il} & \gls{acr:fy} & \gls{acr:glm}, \gls{acr:gnn} & Spanning forest \\
			\cite{parmentierLearningStructuredApproximations2021} & 2SMWST & \gls{acr:il}/\gls{acr:ecm} & \gls{acr:fy} & \gls{acr:glm} & Spanning forest \\
			\bottomrule
		\end{tabular}
		\begin{tablenotes}
			\item Abbreviations: 2SMWST = two-stage stochastic minimum weight spanning tree, 2SIP = two-stage stochastic integer programming, GRU = gated recurrent unit.  
		\end{tablenotes}
	\end{threeparttable}
\end{table}

\paragraph{Hard combinatorial optimization problems.}
A first category (cf.\ Table~\ref{tab:applications_static}) comprises works on hard \gls{acr:co} problems, such as the \gls{acr:svsp}, the \gls{acr:smsp}, or the \gls{acr:fctp}. 
These works typically formulate a tractable surrogate problem, which is then parameterized by a learned statistical model. 
Most contributions employ \gls{acr:il} or \gls{acr:ecm}, using \glspl{acr:glm} or simple \glspl{acr:nn} as encoders and losses such as \gls{acr:fy}, perturbed, or empirical regret. 
Examples include approximate \gls{acr:vsp} heuristics \citep{parmentierLearningApproximateIndustrial2021}, structured heuristics for the \gls{acr:smsp} \citep{parmentier2023structured}, and reduced \gls{acr:fctp} formulations via \glspl{acr:gnn} \citep{spieckermann2025reduce}. 
Together, these works show that \gls{acr:coaml} pipelines can learn scalable heuristics for problems that remain challenging for traditional optimization solvers. 

\paragraph{\gls{acr:cso} problems.}
A second category comprises \gls{acr:cso} problems, where contextual features inform predictions of stochastic parameters. 
Here, the \gls{acr:ml}-layer captures stochasticity, while the \gls{acr:co}-layer guarantees feasibility. 
Representative examples include the \gls{acr:dirp} \citep{greif2024combinatorial}, traffic equilibria prediction \citep{jungel2024wardropnet}, and resource allocation in security games \citep{kong2022end}. 
These architectures frequently combine \gls{acr:il} with \gls{acr:fy}-losses and employ neural models such as \glspl{acr:pinn}, \glspl{acr:nn}, or \glspl{acr:gnn}. 
This category illustrates the promise of \gls{acr:coaml} in bridging predictive modeling and optimization, but also highlights open challenges regarding partial observability and robustness under distribution shift. 

\paragraph{Two-stage stochastic optimization problems.}
A third category concerns two-stage problems, where anticipative first-stage decisions must account for uncertain second-stage outcomes. 
Examples include two-stage stochastic integer programming \citep{bengio2020learning} and two-stage stochastic minimum spanning trees \citep{dalle2022learning}. 
Most contributions here adopt \gls{acr:il} with \gls{acr:fy}-losses, often using \glspl{acr:glm} or \glspl{acr:nn} as encoders. 
These works demonstrate that \gls{acr:coaml} pipelines can extend beyond classical stochastic programming by coordinating first-stage decisions with learned anticipations of second-stage costs. 

\paragraph{Multi-stage stochastic optimization problems.}
Extending beyond two stages yields multi-stage problems (cf.\ Table~\ref{tab:applications_dynamic}), such as dynamic vehicle routing, emergency medical services, or autonomous mobility-on-demand (AMoD). 
\gls{acr:coaml} pipelines typically decompose these problems into a sequence of tractable single-stage surrogates, with \gls{acr:ml}-models providing anticipations of downstream effects. 
Examples include ambulance redeployment \citep{rautenstrauss2025optimization}, \gls{acr:dvrp} \citep{batyCombinatorialOptimizationEnrichedMachine2024}, and AMoD dispatching \citep{jungel2025learning,endersHybridMultiagentDeep2023,hoppeGlobalRewardsMultiAgent2024}. 
While \gls{acr:il} with \gls{acr:fy}-losses is common, \gls{acr:rl} methods are increasingly leveraged, especially in multi-agent settings. 
The main challenge in this category is scalability, as repeated \gls{acr:co}-layer calls can become computationally expensive in large systems and long horizons. 

\paragraph{Multi-stage problems with continuous or large action spaces.}
Finally, some problems involve continuous or very large action spaces, such as robotic control, recommendation systems, or large-scale scheduling. 
Although not combinatorial in the strict sense, these problems pose similar challenges for~\gls{acr:rl}. 
Here, \gls{acr:coaml}-inspired modules constrain or guide exploration in otherwise intractable action spaces. 
Examples include QT-Opt for robotic grasping \citep{kalashnikovQTOptScalableDeep2018}, continuous-action Q-learning \citep{ryuCAQLContinuousAction2020}, and neighborhood-search mappings for large discrete action spaces \citep{dulac-arnoldDeepReinforcementLearning2016,akkermanDynamicNeighborhoodConstruction2024}. 
These contributions show how \gls{acr:coaml} concepts extend to classical \gls{acr:rl}, enabling stable training and scalability in challenging action spaces. 
\begin{table}[!ht]
	\footnotesize
	\centering
	\begin{threeparttable}
		\caption{Representative literature on \gls{acr:coaml} pipelines in dynamic settings}
		\label{tab:applications_dynamic}
		\setlength{\belowcaptionskip}{0.3cm}
		\setlength\tabcolsep{3pt}
		\begin{tabular}{p{2.2cm} p{2.4cm} p{1.4cm} p{3cm} p{2.2cm} p{3cm}}
			\toprule
			\textbf{Reference} & \textbf{Application} & \textbf{Method} & \textbf{Loss function} & \textbf{Stat. model} & \textbf{\gls{acr:co}-layer} \\
			\midrule
			\multicolumn{6}{c}{\textbf{Multi-stage stochastic optimization problems}} \\
			\hline
			\cite{batyCombinatorialOptimizationEnrichedMachine2024} & \gls{acr:dvrp} & \gls{acr:il} & \gls{acr:fy} & \gls{acr:glm} & Prize-collecting VRPTW \\
			\cite{greif2024combinatorial} & \gls{acr:dirp} & \gls{acr:il} & \gls{acr:fy} & \gls{acr:pinn} & Prize-collecting \gls{acr:tsp} \\
			\cite{jungel2025learning} & AMoD dispatch & \gls{acr:il} & \gls{acr:fy} & \gls{acr:glm} & Dispatch/rebalancing \\
			\cite{rautenstrauss2025optimization} & EMS redeployment & \gls{acr:sl} & \gls{acr:fy} & MLP, LR & Matching \\
			\cite{endersHybridMultiagentDeep2023}, \cite{hoppeGlobalRewardsMultiAgent2024}, \cite{woywoodMultiAgentSoftActorCritic2025} & AMoD dispatch (multi-agent) & \gls{acr:rl} & SAC (TD-error) & \gls{acr:nn} & Bipartite matching \\
			\cite{hoppeStructuredReinforcementLearning2025} & \gls{acr:svsp}, DAP, GSPP & \gls{acr:rl} & \gls{acr:fy} & \gls{acr:nn}, \gls{acr:glm} & \gls{acr:vsp}, ranking, shortest path \\
			\cite{yuanReinforcementLearningOptimization2022} & MoD relocation & \gls{acr:rl} & REINFORCE & \gls{acr:nn}, \gls{acr:glm} & Transportation optimization \\
			\cite{liangIntegratedReinforcementLearning2022a}, \cite{xuLargeScaleOrderDispatch2018a}, \cite{linEfficientLargeScaleFleet2018} & MoD dispatch/ relocation & \gls{acr:rl} & Q-learning / TD-error & \gls{acr:nn}, lookup table & Bipartite matching, contextual masking \\
			\hline
			\multicolumn{6}{c}{\textbf{Multi-stage problems with continuous or large action spaces}} \\
			\hline
			\cite{kalashnikovQTOptScalableDeep2018}, \cite{ryuCAQLContinuousAction2020} & Robotic control (grasping) & \gls{acr:rl} & Q-learning (TD-error) & \gls{acr:nn} & Cross-entropy search, MIP \\
			\cite{chowLyapunovbasedApproachSafe2018} & Gridworld, robotics & \gls{acr:rl} & TD-error, policy distillation & \gls{acr:nn} & Lyapunov-based constraints \\
			\cite{dulac-arnoldDeepReinforcementLearning2016}, \cite{wangBiCDDPGBidirectionallyCoordinatedNets2021}, \cite{akkermanDynamicNeighborhoodConstruction2024} & Recommenders, robotics, games, inventory, scheduling & \gls{acr:rl} & DDPG, QAC & \gls{acr:nn} & Neighborhood search (k-nearest neighb., KDTree, simulated annealing) \\
			\cite{vanvuchelenUseContinuousAction2023} & Inventory control & \gls{acr:rl} & PPO & \gls{acr:nn} & Scaling/clipping maskings \\
			\bottomrule
		\end{tabular}
		\begin{tablenotes}
			\item Abbreviations: (A)MoD = (autonomous) mobility-on-demand, EMS = emergency management system, DAP = dynamic assortment problem, GSPP = grid shortest path problem, MLP = multilayer perceptron, LR = linear regression, SAC = soft actor--critic, PPO = proximal policy optimization, DDPG = deep deterministic policy gradient, QAC = Q-actor-critic, TD = temporal difference.  
		\end{tablenotes}
	\end{threeparttable}
\end{table}

\paragraph{Summary.}
Taken together, Tables~\ref{tab:applications_static}--\ref{tab:applications_dynamic} reveal several common features. 
First, \gls{acr:coaml} pipelines are most valuable when problems involve rich combinatorial constraints, large action spaces, or contextual uncertainty. 
Second, structured losses such as \gls{acr:fy} are widely used, as they allow backpropagation through \gls{acr:co}-layers. 
Third, the choice of the statistical model reflects the problem structure: \glspl{acr:glm} are used in linear surrogate settings, \glspl{acr:gnn} in graph-structured problems, and \glspl{acr:pinn} in domains with physical constraints. 
At the same time, studying \gls{acr:cso} under partial observability and multi-stage settings under distributional shift are important directions for future research.

\subsection{Methodological Perspective} \label{sec: state of art methodology}
While Section~\ref{sec: state of art application} reviewed \gls{acr:coaml} pipelines by application domain, it is equally important to analyze them from a methodological standpoint. 
The literature can be organized into three main paradigms: \gls{acr:ecm}, \gls{acr:il}, and \gls{acr:rl}. Table~\ref{tab:methodology_comparison} provides a comparative analysis of the three paradigms.
Table~\ref{table of methodology} summarizes representative contributions in each category, 
\begin{table}[!htp]
	\footnotesize
	\centering
	\begin{threeparttable}
		\caption{Representative literature on \gls{acr:coaml} pipelines from a methodological perspective}
		\label{table of methodology}
		\setlength{\belowcaptionskip}{0.3cm}
		\setlength\tabcolsep{3pt}
		\begin{tabular}{p{2.2cm} p{2.7cm} p{2.2cm} p{2.2cm} p{4.8cm}}
			\toprule
			\textbf{Reference} & \textbf{Learning \mbox{algorithm}} & \textbf{Compatible architecture} & \textbf{Convergence guarantee}  & \textbf{Main challenge addressed}  \\
			\midrule
			\multicolumn{5}{c}{\textbf{Empirical cost minimization (\gls{acr:ecm})}} \\
			\hline
			\cite{aubin2024generalization} & Surrogate policy learning & Stat.\ models (mild assumptions) & Yes (Thm.\ 1) & Piecewise constant risk; generalization guarantees \\
			\cite{bouvier2025primal} & Primal--dual alternating minimization & Deep learning models & Yes (linear conv.) & Non-optimality bounds; scalable \gls{acr:ecm} \\
			\cite{elmachtoubSmartPredictThen2021} & Structured hinge (SPO+) & Linear models & Yes (under assumptions) & Decision-focused surrogate loss \\
			\cite{parmentierLearningStructuredApproximations2021} & Learning heuristics for hard \gls{acr:co} & \gls{acr:glm} & Yes (Thm.\ 2,3) & Bounds on \gls{acr:ecm} error; tractable heuristics \\
			\cite{dalle2022learning} & InferOpt.jl heuristics & \gls{acr:nn}, \gls{acr:glm} & No & Differentiable surrogate optimization \\
			\hline
			\multicolumn{5}{c}{\textbf{Imitation learning (\gls{acr:il})}} \\
			\hline
			\cite{nowozinStructuredLearningPrediction2011} & \gls{acr:sl} with MLE or struct.~Hinge loss & PGM, \gls{acr:glm} & No & Structured learning for applications in vision \\
			\cite{blondelLearningFenchelYoungLosses2020} & \gls{acr:sl} with \gls{acr:fy} losses & \gls{acr:nn} & Yes in \gls{acr:fy}L, No in Emp.~Cost & Unifying perspective on structured learning using \gls{acr:fy}L\\
			\cite{berthet2020learning} & \gls{acr:sl} with perturbation based \gls{acr:fy}L & \gls{acr:nn} & Yes in \gls{acr:fy}L, No in Emp.~cost & Introduce reg.~by perturbation to make \gls{acr:fy}L tractable \\
			\cite{parmentierLearningApproximateIndustrial2021} & Tractable \gls{acr:sl} + CG matheuristic & \gls{acr:nn} & No & Learning surrogate problems to solve hard \gls{acr:co} \\
			\cite{dalle2022learning} & Structured \gls{acr:il} for hard \gls{acr:co} & \gls{acr:nn}, \gls{acr:glm} & No & Differentiable surrogate optimization (InferOpt.jl) \\
			\hline
			\multicolumn{5}{c}{\textbf{Reinforcement learning (\gls{acr:rl})}} \\
			\hline
			\cite{delarueReinforcementLearningCombinatorial2020} & V-learning & \gls{acr:nn} & No & Value learning for combinatorial action spaces \\
			\cite{hoppeStructuredReinforcementLearning2025} & \gls{acr:srl} (primal--dual algorithm) & \gls{acr:nn}, \gls{acr:glm} & Yes (link to mirror descent) & Stable \gls{acr:rl} in combinatorial action spaces \\
			\cite{emamiLearningPermutationsSinkhorn2018} & \gls{acr:rl} with Sinkhorn layers (DDPG-based) & \gls{acr:nn} & No & \gls{acr:rl} for tasks involving permutations (e.g., sorting, \gls{acr:tsp}) \\
			\cite{chowLyapunovbasedApproachSafe2018} & \gls{acr:rl} with Lyapunov functions (distillation) & \gls{acr:nn} & No & Fulfilling safety constraints in \gls{acr:rl} \\
			\cite{caiReinforcementLearningDriven2019} & Policy-based \gls{acr:rl} (PPO) & \gls{acr:nn} & No & Generating initial solutions for heuristics \\
			\cite{yuanReinforcementLearningOptimization2022} & Hybrid of \gls{acr:sl} and \gls{acr:rl} (REINFORCE) & \gls{acr:nn}, \gls{acr:glm} & No & Refining heuristic solutions with \gls{acr:rl} \\
			\cite{linEfficientLargeScaleFleet2018}, \cite{liangIntegratedReinforcementLearning2022a} & Multi-agent Q-learning & \gls{acr:nn} & No & Agent coordination in large action spaces \\
			\cite{endersHybridMultiagentDeep2023} & Multi-agent Soft Actor-Critic & \gls{acr:nn} & No & Agent coordination in large action spaces \\
			\cite{ryuCAQLContinuousAction2020} & Q-learning & \gls{acr:nn} & No & Q-learning in continuous action spaces \\
			\cite{dulac-arnoldDeepReinforcementLearning2016} & \gls{acr:rl} with action search (DDPG) & \gls{acr:nn} & No & Scaling \gls{acr:rl} to large discrete action spaces \\
			\bottomrule
		\end{tabular}
		\begin{tablenotes}
			\item Abbreviations: PGM = Probabilistic Graphical Model.   
		\end{tablenotes}
	\end{threeparttable}
\end{table}
reporting the proposed algorithm, the compatible architectures, whether convergence guarantees are available, and the main challenge addressed. 

\paragraph{Empirical cost minimization.}
In \gls{acr:ecm}, the learner minimizes empirical cost directly using observed samples $(x_i,\xi_i)$. 
The main advantage is its strict decision focus, as performance is measured on the actual combinatorial task rather than on prediction accuracy. 
An immediate advantage is that it enables to give performance guarantees that include statistical generalization~\citep{aubin2024generalization}.
Since the empirical cost is piecewise constant in the model parameters, \citet{parmentierLearningStructuredApproximations2021} suggests regularizing it and minimizing it using a global optimization algorithm such as Bayesian optimization.
While this works when the parameter $w$ is small dimensional, which happens for instance when $\varphi_w$ is a \gls{acr:glm}, it prevents the use of deep learning.
To train deep architectures, \citet{dalle2022learning}  regularize the problem and use reinforce gradients, i.e., a score-based estimator~\citep{blondel_edpbook}: while it has the advantage of simplicity, performance suffers from non-convexity and the noisy gradients. 
The papers listed in Table~\ref{table of methodology} address this through surrogate losses, e.g., SPO+ \citep{elmachtoubSmartPredictThen2021}, or convex reformulations minimized using a primal--dual alternating minimization scheme \citep{bouvier2025primal}. 
Both lead to good performances for learning deep architectures, but come with limitations: the former require a linear objective, while the latter require a single scenario optimization oracle.

\paragraph{Imitation learning.}
\gls{acr:il} leverages expert decisions $y^\ast(x)$, often obtained by solving deterministic surrogate problems, and trains a statistical model to reproduce them, typically using a \gls{acr:fy}-loss~\citep{blondelLearningFenchelYoungLosses2020,berthet2020learning}. 
This paradigm has been applied extensively to problems such as the \gls{acr:svsp} and the \gls{acr:smsp} \citep{parmentierLearningApproximateIndustrial2021,parmentierLearningStructuredApproximations2021}, and more recently to multi-stage problems such as the \gls{acr:dvrp} \citep{batyCombinatorialOptimizationEnrichedMachine2024}. 
The main strength of \gls{acr:il} is its practicality: efficient solvers can generate abundant training data without requiring access to the full distribution of $\bfxi$. 
However, the quality of the learned policy is inherently bounded by that of the expert solutions, and biases in the surrogate propagate into the learned model. 
Recent works attempt to overcome this by combining \gls{acr:il} with decision-focused, e.g., by integrating \gls{acr:ecm}-style updates, but systematic guarantees remain limited.

\paragraph{Reinforcement learning.}
\gls{acr:rl}-based \gls{acr:coaml} is most relevant when dynamics are implicit or only accessible via simulation. 
Here, policies are trained from interaction data, and rewards guide the optimization process. 
Classical \gls{acr:rl} methods struggle with combinatorial action spaces, which motivates embedding \gls{acr:co}-layers into policies. 
This ensures feasibility by construction and constrains exploration to meaningful parts of the action space. 
\gls{acr:srl} is a recent example, integrating \gls{acr:co}-layers into actor architectures and \gls{acr:fy}-losses into actor-critic updates to stabilize training \citep{hoppeStructuredReinforcementLearning2025}. 
Other contributions (cf.\ Table~\ref{table of methodology}) focus on hybrid \gls{acr:rl} architectures for mobility-on-demand, 
robotics, or recommender systems \citep{endersHybridMultiagentDeep2023,kalashnikovQTOptScalableDeep2018,dulac-arnoldDeepReinforcementLearning2016}. 
The main methodological challenge is computational: actor updates often require multiple calls to the \gls{acr:co}-layer per iteration, which can dominate training cost.

\paragraph{Synthesis.}
Taken together, Tables~\ref{tab:methodology_comparison}\&\ref{table of methodology} show how \gls{acr:ecm}, \gls{acr:il}, and \gls{acr:rl} represent three complementary paradigms. 
\gls{acr:ecm} offers theoretical rigor but limited scalability, \gls{acr:il} provides practical effectiveness but depends on the quality of expert data, and \gls{acr:rl} enables learning under implicit dynamics but at high computational cost. 
Their intersection reveals common methodological challenges: propagating informative gradients through non-differentiable \gls{acr:co}-layers, scaling to deep models, and establishing convergence guarantees. 
Future progress in \gls{acr:coaml} is likely to emerge from approaches that blend these paradigms, combining the decision-focus of \gls{acr:ecm}, the practicality of \gls{acr:il}, and the flexibility of~\gls{acr:rl}.

\subsection{Statistical Learning Guarantees}\label{subsec:guarantees}
Beyond requiring no target $\bar y$ in the training set, one advantage of risk minimization is that it provides guarantees on the solution returned.
Let us denote by $R^* = \min_{\pi \in \calH}\calR(h)$ the risk of the best policy where $\calH$ contains all possible policies (not necessarily based on \gls{acr:coaml}), and by $R^\dagger = \min_{w}\calR(\pi_w)$ the risk of the best policy in the class $\calH_{\calW}$. The gap
$$R^* - R^\dagger \geq 0 $$
is the \emph{model bias} and can only be improved by changing the class $\calH_{\calW}$.
\citet{demelas2024predicting} provide a first attempt that exploits deep learning and Lagrangian duality to bound this model bias.

The goal of the learning algorithm is to find a $w$ such that $\calR(\pi_w)$ is as close as possible to $R^\dagger$.
\citet{aubin2024generalization} prove an upper bound on the excess risk $\calR(\hat w_n)- R^\dagger$ of the $\hat w_n$ returned by their learning algorithm that stands with high probability on the drawing of the training set $x_1,\ldots,x_n$.
They also show that $\calR(\hat w_n)- R^\dagger$ converges to $0$ with the size of the training set.
\citet{parmentierLearningStructuredApproximations2021} shows that on some problems, the model bias can be bounded, and thus $\pi_{\hat w_n}$ is an approximation algorithm where the approximation ratio is controlled in expectation over~$\bbP_X$.

In the specific setting of linear objectives, \citet{elmachtoubSmartPredictThen2021} provide rigorous regret bounds for the SPO+ loss, showing that it is Fisher consistent with the true regret.
More recently, \citet{capitaine2025} further analyze this setting, deriving explicit non-regret bounds for linear objectives.
However, these theoretical guarantees rely fundamentally on the linearity of the cost function.
Consequently, extending these consistency and regret analyses to the general non-linear and non-convex \gls{acr:coaml} architectures remains a significant open challenge in the field.
\section{Conclusion \& Perspectives}
\Gls{acr:coaml} has rapidly developed 
into a promising paradigm at the interface of \gls{acr:or} and \gls{acr:ml}. 
The central idea is to embed \gls{acr:co} oracles within learning pipelines, 
thereby combining the predictive power of statistical models with the feasibility 
guarantees and structural reasoning of optimization. 
This survey has aimed to provide a comprehensive and tutorial-style overview of 
this emerging field, covering its technical foundations, methodological advances, 
and application domains. 

In contrast to broader surveys of contextual decision-making and decision-focused learning 
\citep[e.g.,][]{sadana2024survey, mandiDecisionFocusedLearningFoundations2024, misicDataAnalyticsOperations2020}, 
our work offers a focused treatment of \gls{acr:coaml}. 
It contributes (i) a unified formalization that situates \gls{acr:coaml} relative to classical \gls{acr:co}, 
\gls{acr:pto}, and structured prediction, 
(ii) a detailed development of \gls{acr:ecm} as a conceptual foundation, 
and (iii) coverage of \gls{acr:srl}, an aspect largely absent from 
earlier reviews. We formalized \gls{acr:coaml} pipelines and situated them within the broader landscape of structured prediction, predict-then-optimize, and decision-focused learning. 
Section~3 discussed the main methodological building blocks, with particular emphasis on \gls{acr:ecm}, which provides the conceptual foundation for aligning prediction and decision-making and remains the source of many open challenges. 
Section~4 offered a structured overview of the state of the art: it introduced a taxonomy of problem settings (explicit vs.\ implicit uncertainty, static vs.\ dynamic problems), reviewed algorithms for static and dynamic cases, analyzed applications across five major categories, and synthesized methodological paradigms in \gls{acr:ecm}, \gls{acr:il}, and \gls{acr:rl}. 
Taken together, these perspectives highlight both the breadth of \gls{acr:coaml} and the importance of aligning algorithmic choices with problem characteristics. 

Despite significant progress, several recurring challenges remain unresolved. 
First, \emph{gradient propagation} through non-differentiable \gls{acr:co}-layers remains difficult, often requiring surrogate losses or perturbation-based methods that introduce bias or variance. 
Second, \emph{scalability} is a central bottleneck: many pipelines require repeated calls to \gls{acr:co}-oracles during training, which can be computationally prohibitive in large-scale or multi-stage settings. 
Third, \emph{generalization} under distributional shift, partial observability, or scarce data remains poorly understood; most current methods assume that training and deployment distributions coincide. 
Fourth, the lack of \emph{standard benchmarks and evaluation protocols} makes it difficult to compare methods across studies, hindering reproducibility and cumulative progress. 
These challenges cut across paradigms and represent fundamental obstacles to wider adoption of \gls{acr:coaml}. 

Looking forward, we identify several key methodological challenges and corresponding research directions that constitute a coherent agenda for advancing \gls{acr:coaml}, summarized in Table~\ref{tab:research-agenda}. First, to address the challenge of efficient exploration in high-dimensional spaces where combinatorial oracles are computationally prohibitive, we envision the development of \emph{hybrid paradigms}. These should combine the decision focus of \gls{acr:ecm}, the practicality of \gls{acr:il}, and the flexibility of \gls{acr:rl} to balance theoretical guarantees with empirical effectiveness. Second, there is a pressing need for structured architectures capable of integrating sequential information with combinatorial reasoning. Future work should align architectures with problem structures, utilizing graph \glspl{acr:nn} for networked systems, autoregressive models for sequential decision-making, or physics-informed models for domains with conservation laws. Third, distinct from standard fully observable settings, many practical applications involve noisy or indirect data. A major frontier is therefore the integration of uncertainty modeling, enabling pipelines to handle implicit distributions, partial observability, and multi-stage stochasticity within a unified framework. Finally, to ensure generalization across environments, empirical advances must be matched with theoretical rigor. This includes establishing stronger convergence guarantees for deep \gls{acr:coaml} pipelines and gaining complexity-theoretic insights into which problem classes admit efficient decision-focused learning. Supporting this agenda requires the creation of benchmarks and open-source toolkits for systematic comparison, as well as validation through cross-disciplinary applications in logistics, energy, healthcare, and robotics.

While this survey has focused on \gls{acr:co}-layers, 
similar challenges arise in the closely related field of 
\emph{continuous optimization layers}. 
Here, differentiable convex optimization modules 
\citep[e.g.,][]{amos2017,agrawal2019differentiable} and KKT-based implicit layers provide tractable ways of 
embedding optimization problems into end-to-end pipelines. 
Other approaches rely on unrolled optimization or fixed-point methods 
to enable gradient propagation through continuous solvers. 
Although the techniques differ from those in combinatorial settings, 
the core questions are analogous: how to propagate gradients reliably, 
how to balance tractability with accuracy, and how to ensure generalization 
under distributional shift. 
Future research may benefit from cross-fertilization between the two fields, 
as ideas developed for continuous layers could inspire 
novel architectures and training methods for \gls{acr:coaml}.

\gls{acr:coaml} is still in its early stages, but its potential to transform decision-making under uncertainty is considerable. 
By uniting the traditions of \gls{acr:or} and \gls{acr:ml}, the field has the opportunity to produce algorithms that are both theoretically grounded and practically applicable. 
Future progress will require bridging paradigms, advancing architectures and losses, developing systematic benchmarks, and deepening theoretical understanding. 
Placing \emph{\gls{acr:ecm} at the core} of this agenda is particularly important, as it is the unifying framework that enables \gls{acr:coaml} to connect predictive modeling, optimization, and generalization theory. 
If these challenges can be addressed, \gls{acr:coaml} may become a cornerstone paradigm for intelligent decision-making in complex, uncertain, and combinatorial environments. 
We hope this survey provides a foundation for researchers and practitioners alike, clarifying the state of the art and pointing to the rich set of research frontiers that lie ahead.
\begin{table}[!hb]
	\centering
	\footnotesize
	\caption{Research agenda for \gls{acr:coaml}: key frontiers and directions.}
	\label{tab:research-agenda}
	\begin{tabular}{p{2.7cm}p{14cm}}
		\toprule
		\textbf{Frontier} & \textbf{Research Directions} \\
		\midrule
		\textbf{Hybrid Paradigms \& Exploration} & Develop hybrid frameworks combining the decision focus of \gls{acr:ecm}, the practicality of \gls{acr:il}, and the flexibility of \gls{acr:rl}; address efficient exploration in high-dimensional spaces to reduce prohibitive oracle calls; design surrogate optimization layers. \\
		\addlinespace[0.4em]
		\textbf{Structured Architectures} & Design architectures that align with combinatorial structures: graph \glspl{acr:nn} for network problems, autoregressive models for sequential data, and physics-informed layers; integrate attention mechanisms with combinatorial reasoning. \\
		\addlinespace[0.4em]
		\textbf{Uncertainty \& Observability} & Move beyond full observability to handle partial observability and noisy contexts; integrate implicit distributions and uncertainty quantification; unify robust and stochastic optimization within \gls{acr:coaml} pipelines. \\
		\addlinespace[0.4em]
		\textbf{Theory \& Generalization} & Establish convergence guarantees for deep \gls{acr:coaml} pipelines; analyze generalization bounds across distribution shifts (environment adaptation); derive complexity-theoretic insights into decision-focused learning. \\
		\addlinespace[0.4em]
		\textbf{Benchmarks \& \mbox{Reproducibility}} & Create standardized problem generators, datasets, and open-source toolkits; enable systematic comparison of algorithms across diverse domains; foster reproducible empirical evaluation. \\
		\addlinespace[0.4em]
		\textbf{Applications} & Validate methods in high-stakes domains like logistics, energy, healthcare, and robotics; leverage cross-disciplinary constraints to expose methodological gaps and motivate new algorithmic formulations. \\
		\bottomrule
	\end{tabular}
\end{table}

\FloatBarrier
\newpage

\appendix
\section{Overview of used Acronyms}

\renewcommand{\glossarysection}[2][]{}
\printnoidxglossaries

\let\oldthebibliography\thebibliography
\renewcommand\thebibliography[1]{%
  \oldthebibliography{#1}
  \setlength{\itemsep}{4pt}
}
\bibliographystyle{plainnat}
\bibliography{biblio}

\end{document}